\documentclass{article}

\usepackage{microtype}
\usepackage{graphicx}
\usepackage{subcaption}
\usepackage{booktabs} 
\usepackage{adjustbox}
\usepackage{multirow}
\usepackage{amsmath}
\usepackage{amsthm}
\usepackage{amssymb}
\usepackage{graphicx}
\usepackage{wrapfig}
\usepackage{xspace}
\usepackage{extarrows}
\usepackage{xfrac}
\usepackage{tcolorbox}
\usepackage{sidecap}        \sidecaptionvpos{figure}{t}
\usepackage{enumitem}
\setitemize{noitemsep,topsep=0pt,parsep=0pt,partopsep=0pt}

\renewcommand{\vec}[1]{\mathbf{#1}}
\newcommand{\mat}[1]{\mathbf{#1}}
\newcommand{\x}[0]{\vec{x}}

\newcommand{\z}[0]{\vec{z}}
\newcommand{\J}[0]{\mat{J}}

\newcommand{\dif}[1]{\mathrm{d}#1} 
\newcommand{\KL}[2]{\text{KL}({#1}||{#2})} 
\newcommand{\T}[0]{^{\top}}

\def\R{\mathbb{R}}  
\def\E{\mathbb{E}}  
\def\N{\mathcal{N}} 
\def\M{\mathcal{M}} 
\def\Z{\mathcal{Z}} 
\def\X{\mathcal{X}} 

\newtheorem{definition}{Definition}
\theoremstyle{definition}

\newtcolorbox{exbox}{colback=gray!15,colframe=black}

\numberwithin{equation}{section}

\makeatletter
\newcommand*{\etc}{%
	\@ifnextchar{.}%
	{etc}%
	{etc.\@\xspace}%
}
\makeatother
\usepackage{hyperref}

\usepackage[accepted]{icml2020}

\icmltitlerunning{Variational Autoencoders with Riemannian Brownian Motion Priors}

\begin{document}
	
	\twocolumn[
	\icmltitle{Variational Autoencoders with Riemannian Brownian Motion Priors}
	
	
	
	\icmlsetsymbol{equal}{*}
	
	\begin{icmlauthorlist}
		\icmlauthor{Dimitris Kalatzis}{dtu}
		\icmlauthor{David Eklund}{rise}
		\icmlauthor{Georgios Arvanitidis}{mp}
		\icmlauthor{S{\o}ren Hauberg}{dtu}
	\end{icmlauthorlist}
	
	\icmlaffiliation{dtu}{Section for Cognitive Systems, Department of Applied Mathematics and Computer Science, Technical University of Denmark}
	\icmlaffiliation{mp}{Empirical Inference Department, Max Planck Institute for Intelligent Systems, T\"{u}bingen, Germany}
	\icmlaffiliation{rise}{Research Institutes of Sweden, Isafjordsgatan 22, 164 40 Kista, Sweden}
	
	\icmlcorrespondingauthor{Dimitris Kalatzis}{dika@dtu.dk}
	
	\icmlkeywords{VAE, Differential Geometry, Riemannian Geometry}
	
	\vskip 0.3in
	]
	
	
	
	\printAffiliationsAndNotice{}  
	
	\begin{abstract}
		Variational Autoencoders (VAEs) represent the given data in a low-dimensional latent space, which is generally assumed to be Euclidean. This assumption naturally leads to the common choice of a standard Gaussian prior over continuous latent variables. Recent work has, however, shown that this prior has a detrimental effect on model capacity, leading to subpar performance. We propose that the Euclidean assumption lies at the heart of this failure mode. To counter this, we assume a Riemannian structure over the latent space, which constitutes a more principled geometric view of the latent codes, and replace the standard Gaussian prior with a Riemannian Brownian motion prior. We propose an efficient inference scheme that does not rely on the unknown normalizing factor of this prior. Finally, we demonstrate that this prior significantly increases model capacity using only one additional scalar parameter.
	\end{abstract}
	
	\section{Introduction}
	Variational autoencoders (VAEs) \citep{kingma:iclr:2014, rezende2014stochastic} simultaneously learn a conditional density $p(\x | \z)$ of high dimensional observations and low dimensional representations $\z$ giving rise to these observations. In VAEs, a prior distribution $p(\z)$ is assigned to the latent variables which is typically a standard Gaussian. It has, unfortunately, turned out that this choice of distribution is limiting the modelling capacity of VAEs and richer priors have been proposed instead \cite{tomczak2017vae, van2017neural, bauer2018resampled, klushyn2019learning}. In contrast to this popular view, we will argue that the limitations of the prior are not due to \emph{lack of capacity}, but rather \emph{lack of principle}.
	
	\begin{figure}
		\centering
		\includegraphics[height=2.8cm]{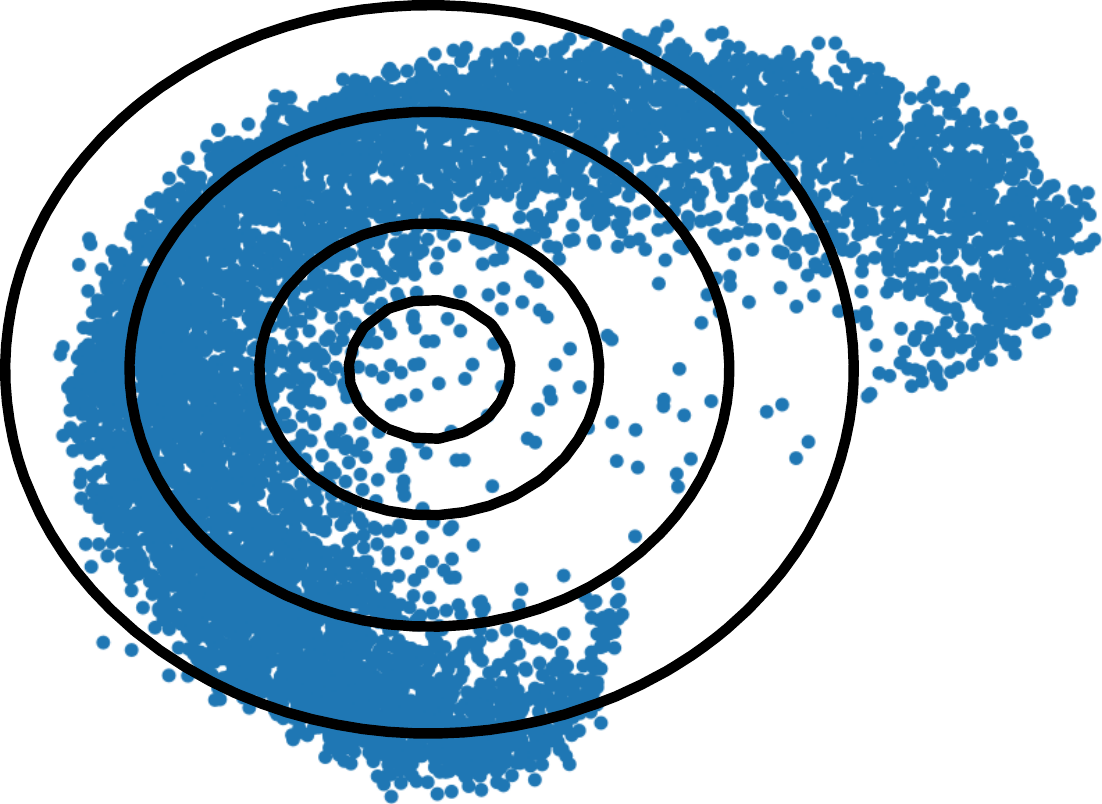}
		\includegraphics[height=2.8cm]{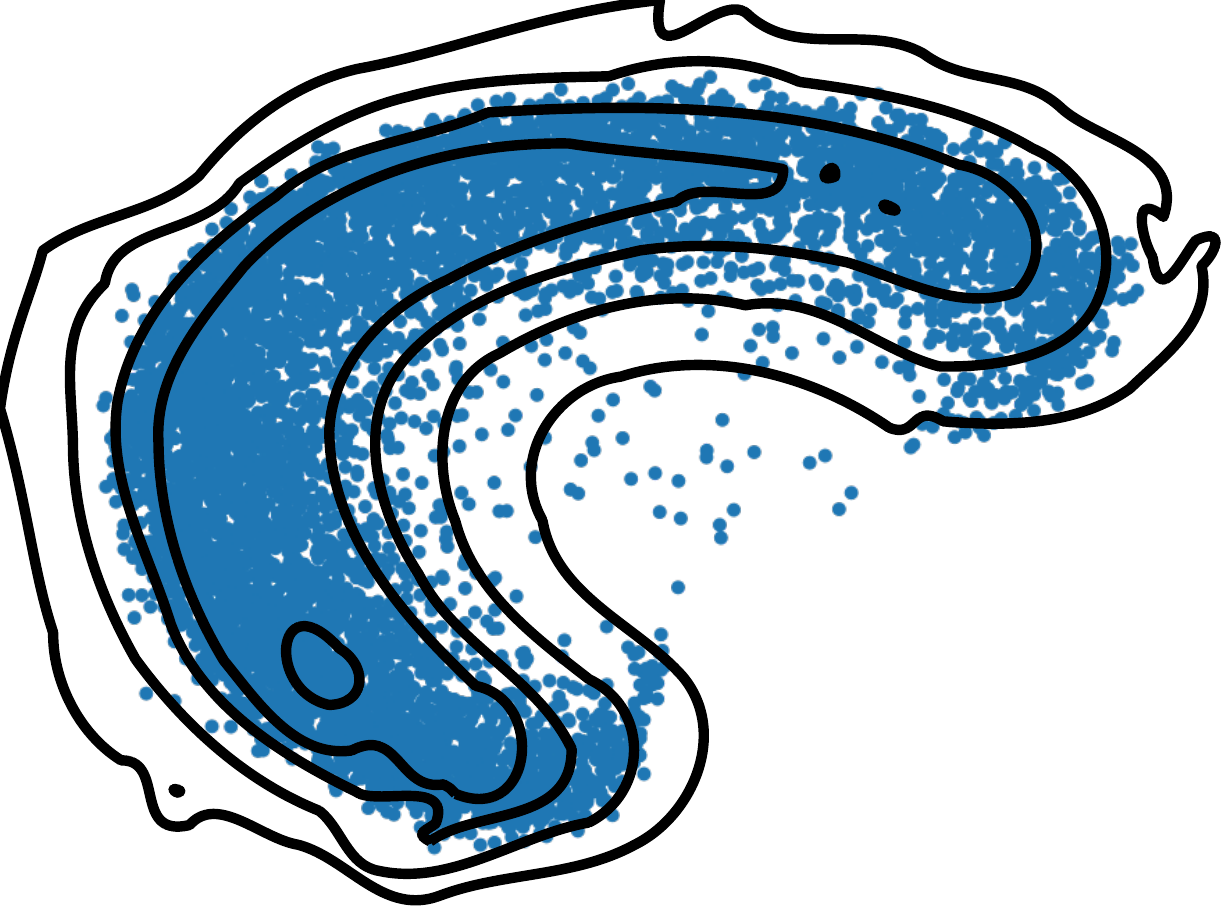}
		\caption{The latent space priors of two VAEs trained on the digit \textit{1} from MNIST. \emph{Left:} Using a unit Gaussian prior. \emph{Right:} Using a Riemannian Brownian motion (ours) with trainable (scalar) variance.}
		\label{fig:mnist_teaser}
	\end{figure}
	
	Informally, the Gaussian prior has two key problems.
	
	\textbf{1. $\quad$ The Euclidean representation is arbitrary.}
	Behind the Gaussian prior lies the assumption that the latent space $\mathcal{Z}$ is Euclidean.
	However, if the decoder $p_\theta(\x|\z)$ is of sufficiently high capacity, then it is always possible to reparameterize the latent space from $\z$ to $h(\z), h: \mathcal{Z} \rightarrow \mathcal{Z}$, and then let the decoder invert this reparameterization as part of its decoding process \citep{arvanitidis:iclr:2018, hauberg2018only}. This implies that we cannot assign any meaning to specific instantiations of the latent variables, and that Euclidean distances carry limited meaning in $\mathcal{Z}$. This is an \emph{identifiability problem} and it is well-known that even the most elementary latent variable models are subject to such. For example, Gaussian mixtures can be reparameterized by permuting cluster indices, and principal components can be arbitrarily rotated \citep{Bishop:2006:PRM:1162264}.
	
	\textbf{2.$\quad$ Latent manifolds are mismapped onto $\boldsymbol{\mathcal{Z}}$}.
	In all but the simplest cases, the latent manifold $\mathcal{M}$ giving rise to data observations is embedded in $\mathcal{Z}$. An encoder with adequate capacity will always recover some smoothened form of $\mathcal{M}$, which will either result in the latent space containing ``holes'' of low density or, in $\mathcal{M}$ being mapped to the whole of $\mathcal{Z}$ under the influence of the prior. Both cases will lead to bad samples or convergence problems. This problem is called \emph{manifold mismatch} \cite{davidson2018hyperspherical, falorsi2018explorations} and is closely related to \emph{distribution mismatch} \cite{hoffman2016elbo, bauer2018resampled, rosca2018distribution} where the prior samples from regions to which the variational posterior (or encoder) does not assign any density. A graphical illustration of this situation can be seen on the left panel of Fig.~\ref{fig:mnist_teaser}, where a VAE is trained on the 1-digits of MNIST under the Gaussian prior. The prior assigns density where there is none.
	
	\textbf{In this paper}, we consider an alternative prior, which is shown in the right panel of Fig.~\ref{fig:mnist_teaser}. This is a Riemannian Brownian motion model defined over the manifold immersed by the decoder. The Riemannian structure solves the identifiability problem and gives a meaningful representation that is invariant to reparametrizations and at the same time restricts the prior to sample only from the image of $\mathcal{M}$ onto $\mathcal{Z}$. The prior generalizes the Gaussian to the Riemannian setting. It only has a single scalar variance parameter, yet it is able to capture intrinsic complexities in the data.
	
	\section{Background}
	\subsection{Variational autoencoders}
	VAEs learn a generative model $p_\theta(\vec{x, z})$ by specifying a likelihood of observations conditioned on latent variables $p_\theta(\vec{x|z)}$ and a prior over the latent variables $p(\z)$. The marginal likelihood of the observations $p_\theta(\x) = \int p_\theta(\vec{x | z}) p(\vec{z}) \mathrm{d}\vec{z}$ is intractable. As such, VAEs are trained by maximizing the variational \emph{Evidence Lower Bound} (ELBO) on the marginal likelihood : 
	\begin{align}
	\E_{q(\vec{z}|\vec{x})}[\log p_\theta(\vec{x|z})] - \KL{q_\phi(\z|\x)}{p(\z)}, \label{eq:ELBO}
	\end{align}
	where $q_\phi(\z|\x)$ denotes the variational family. \citet{kingma:iclr:2014, rezende2014stochastic} proposed a low variance estimator of stochastic gradients of the ELBO, known as \emph{reparameterization trick}.
	
	In the VAE framework, both the variational family $q_\phi(\z|\x)$ and the conditional likelihood $p_\theta(\x|\z)$ are parameterized by neural networks with variational parameters $\phi$ and generative parameters $\theta$. In the language of autoencoders, these networks are often called \emph{encoder} and \emph{decoder} parameterizing the variational family and the generative model respectively. From an autoencoder perspective, Eq.~\ref{eq:ELBO} can be seen as a loss function involving a data reconstruction term (the generative model) and a regularization term (the KL divergence between the variational family and the prior distribution over the latent variables).
	
	\subsection{A primer on Riemannian geometry} \label{review}
	The standard Gaussian prior relies on the usual Lebesgue measure which in turn, assumes a Euclidean structure over the latent space $\mathcal{Z}$. Recently, it has been noted \cite{arvanitidis:iclr:2018, hauberg2018only} that this assumption is mathematically questionable, and that, empirically, Euclidean latent space distances carry little information about the relationship between data points. Rather, a Riemannian interpretation of the latent space appears more promising. 
	Hence we give a short review of Riemannian geometry.
	
	A \textit{smooth manifold} $\M$ is a topological manifold endowed with a smooth structure. That is to say $\M$ is locally homeomorphic to Euclidean space and we are able to do calculus on it. For a point $p \in \M$, the \textit{tangent space} $T_p\M$ is a vector space centered on $p$ which contains all tangent vectors to $\M$ passing through point $p$. With this we can give a formal definition of the Riemannian metric tensor which is of central importance to any analysis involving Riemannian geometry.
	
	\begin{definition} \textbf{(Riemannian metric)} \cite{docarmo:1992}
		Given a smooth manifold $\mathcal{M}$, a Riemannian metric on $\mathcal{M}$ assigns on each point $p \in \mathcal{M}$ an inner product (i.e. a symmetric, positive definite, bilinear form) $\langle\cdot,\cdot\rangle_p$ in the tangent space $T_p\mathcal{M}$ which varies smoothly in the following sense: if $\mathbf{x}: \mathbb{R}^n \supset U \rightarrow \mathcal{M}$ is a local coordinate chart centered at $p$ and $\frac{\partial}{\partial x_{i}}(q)= \mathbf{dx}_{q}(0, \ldots, 1, \ldots, 0)$ for $q \in U$, then $\langle\frac{\partial}{\partial x_{i}}(q), \frac{\partial}{\partial x_{j}}(q)\rangle_{\mathbf{x}(q)} = g_{i j}(q)$ is a smooth function on $U$.\looseness=-1
	\end{definition}
	
	By generalizing the inner product to Riemannian manifolds, the metric tensor gives meaning to length, angle and volume on manifolds. Central to distributions defined on a Riemannian manifold, the volume measure over an infinitesimal region centered at point $p$ is defined as $d\M_p = \sqrt{\mathbf{\det G_p}} \dif p$, where $\mat{G_p}$ is the matrix representation of the metric tensor evaluated at point $p$. Shortest paths on manifolds are represented by geodesic curves, which generalize straight lines in Euclidean space. A geodesic is a constant speed curve and its length can be computed by integrating the norm of its velocity vector under the metric, in other words $\mathcal{L} = \int_{0}^{1} ||\frac{\dif \gamma}{\dif t}||_g \dif t$. For $p \in \M$ there is a useful map defined on a neighborhood of the origin of $T_p\M$ called the exponential map. More precisely, the exponential map is a diffeomorphism, i.e. a bijection with a smooth inverse, between an open subset $\mathcal{U} \subset T_p\M$ and an open subset $\mathcal{U}' \subset \M$. Given $p \in \M$ and $v \in \mathcal{U}$, there is a unique geodesic $\gamma:[0,1] \to \M$ with $\gamma(0)=p$ and $\frac{\dif \gamma}{\dif t}(0)=v$. The exponential map is given by $exp_p(v)=\gamma(1)$. Note that $exp_p(0) = p$. The inverse map (from $\mathcal{U}'$ to $\mathcal{U}$) $exp_p^{-1} = log_p$ is called the \textit{logarithmic map}.
	
	Let $\mathcal{M} \subseteq \mathbb{R}^M$ be an embedded $n$-dimensional manifold and consider local coordinates $\phi:U \to \mathcal{M}$ with $U \subseteq \mathbb{R}^N$ an open subset. The Euclidean metric on $\mathbb{R}^M$ induces a Riemannian metric on $\mathcal{M}$. Expressed in terms of the coordinates given by $\phi$, this metric is known as the \emph{pull-back metric} on $U$ under $\phi$. For $\mathbf{u} \in U$, the pull-back metric $\mathbf{G}_\mathbf{u}$ at $\mathbf{u}$ is given by 
	\begin{align}
	\mathbf{G}_\mathbf{u} = \mat{J}_\phi^{\top}(\mathbf{u})\mat{J}_\phi(\mathbf{u}),
	\end{align}
	where $\mat{J}_\phi$ denotes the Jacobian matrix of $\phi$.
	
	\subsection{VAE decoders as immersions} \label{immersions}
	We will dedicate this subsection to showing that, under certain architectural choices, VAE decoders induce Riemannian metrics in the latent space. That is to say, they belong to a certain class of maps, called \textit{smooth immersions}, which give rise to \textit{immersed submanifolds}. In other words, we will formally describe our intuition about VAEs mapping the latent space back to data space, using the language of smooth manifolds and Riemannian geometry.
	
	The generative and variational distributions can be seen as families of parameterized mappings $g_\phi: \mathcal{X} \rightarrow \mathcal{Z}$ and $f_\theta:\mathcal{Z} \rightarrow \mathbb{R}^M$, $\mathcal{Z} \subset \mathbb{R}^N$ and $M > N$ and parameters $\phi$ and $\theta$ respectively.  The family defined by the generative model is of particular interest. To make the subsequent exposition clearer we will assume a Gaussian generative model and rewrite it in the following form:
	\begin{align} \label{eq:f-map}
	f_\theta(\mathbf{z})=\mu_{\theta}(\mathbf{z})+\sigma_{\theta}(\mathbf{z}) \odot \epsilon, \quad \epsilon \sim \mathcal{N}\left(\mathbf{0}, \mathbb{I}_{M}\right)
	\end{align}
	with $\mu_{\theta}: \mathcal{Z} \rightarrow \mathbb{R}^M$, $\sigma_{\theta}: \mathcal{Z} \rightarrow \mathbb{R}_{+}^{M}$, denoting the mean and standard deviation of the generative model parameterized by neural networks with parameters $\theta$ and $\odot$ denoting the Hadamard or element-wise product. 
	
	\begin{definition} \textbf{(Smooth immersions)} \label{def:immersions}
		Given smooth manifolds $\mathcal{M}$ and $\mathcal{M}'$ with dim($\mathcal{M}$) $<$ dim($\mathcal{M}'$), a mapping $f:\mathcal{M} \rightarrow \mathcal{M}'$, a point $p \in \mathcal{M}$ and its image $f(p) \in \mathcal{M}'$, the mapping $f$ is called an immersion if its differential $\dif f_{p}: T_p\mathcal{M} \rightarrow T_{f(p)}\mathcal{M}'$ is injective for all $p \in \mathcal{M}$. 
	\end{definition}
	
	We will consider a particular Riemannian metric on $\Z$ induced by $\mu_{\theta}$ and $\sigma_{\theta}$. The architectures of $\mu_{\theta}$ and $\sigma_{\theta}$ are such that these maps are immersions. Consider now the \emph{diagonal immersion} 
	\begin{align} f:\Z \to \R^M \times \R_{+}^{M}:\z \mapsto (\mu_{\theta}(\z), \sigma_{\theta}(\z)),
	\end{align} whose geometry encodes both mean and variance. The random map $f_{\theta}$ is a random projection given by $\epsilon$ of the diagonal immersion. Sampling using the decoder can therefore be seen as first sampling the image of this immersion and then randomly projecting down to $\X$ \cite{haubergeklund2019}. Taking the pull-back metric $\mathbf{G_z}$ of $f$ to $\Z$ we obtain 
	\begin{align}
	\mathbf{G_z}=\mat{J_{\boldsymbol{\mu}}}(\vec{z})^{\top} \mat{J_{\boldsymbol{\mu}}}(\vec{z}) + \mat{J_{\boldsymbol{\sigma}}}(\vec{z})^{\top} \mat{J_{\boldsymbol{\sigma}}}(\vec{z}),
	\label{eq:pullback}
	\end{align} where $\mat{J_{\boldsymbol{\mu}}}$ and $\mat{J_{\boldsymbol{\sigma}}}$ are the Jacobian matrices of $\mu_{\theta}$ and $\sigma_{\theta}$.
	
\begin{figure}
		\centering
		\includegraphics[width=0.5\columnwidth]{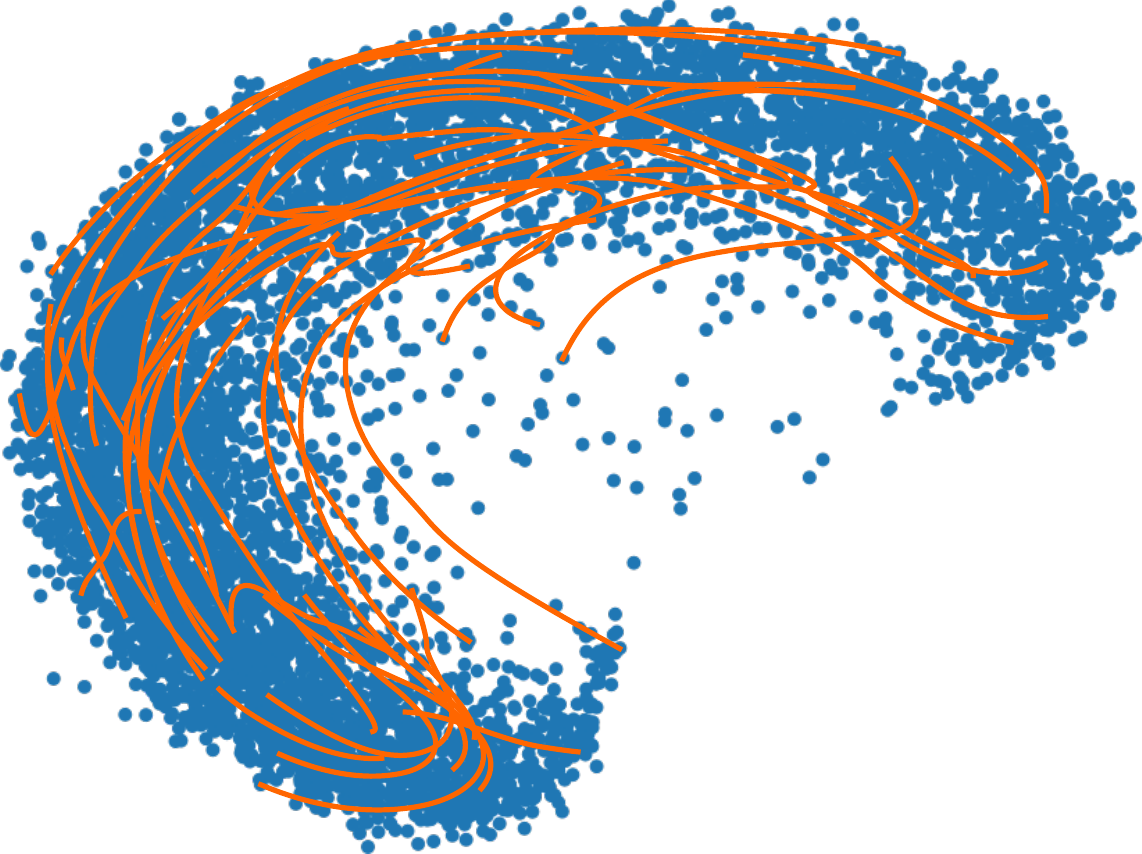}
		\caption{Example geodesics under the pull-back metric \eqref{eq:pullback}. The associated VAE is the same as in Fig.~\ref{fig:mnist_teaser}.}
		\label{fig:geodesics}
	\end{figure}

	The metric $\mathbf{G_z}$ was studied by \citet{arvanitidis:iclr:2018} and is known to yield geodesics that follow high density regions in latent space. As an example, Fig.~\ref{fig:geodesics} shows geodesics of a VAE trained on 1-digits from MNIST, which follow the data due to the variance term of the metric, which penalizes geodesics going through low density regions of the latent space.
	
	\begin{figure*}[t]
		\centering
		\includegraphics[width=0.23\textwidth]{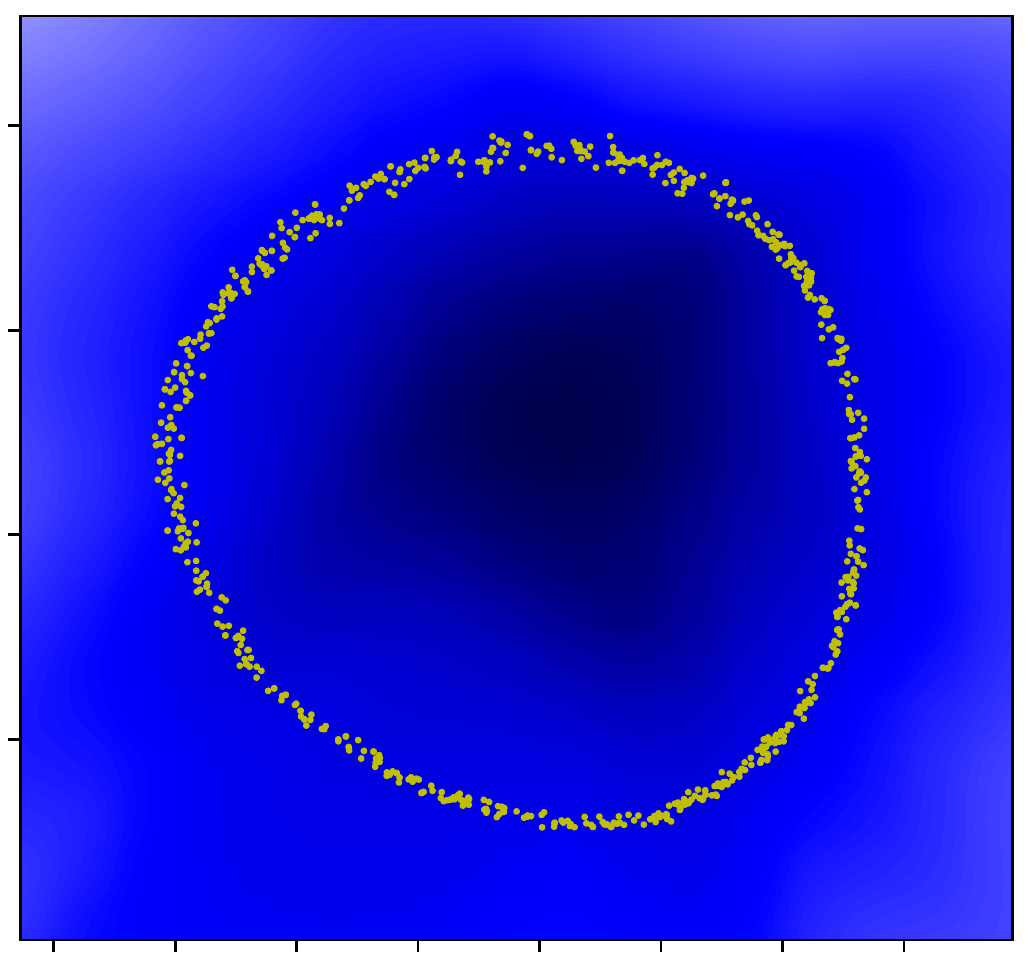}
		\includegraphics[width=0.23\textwidth]{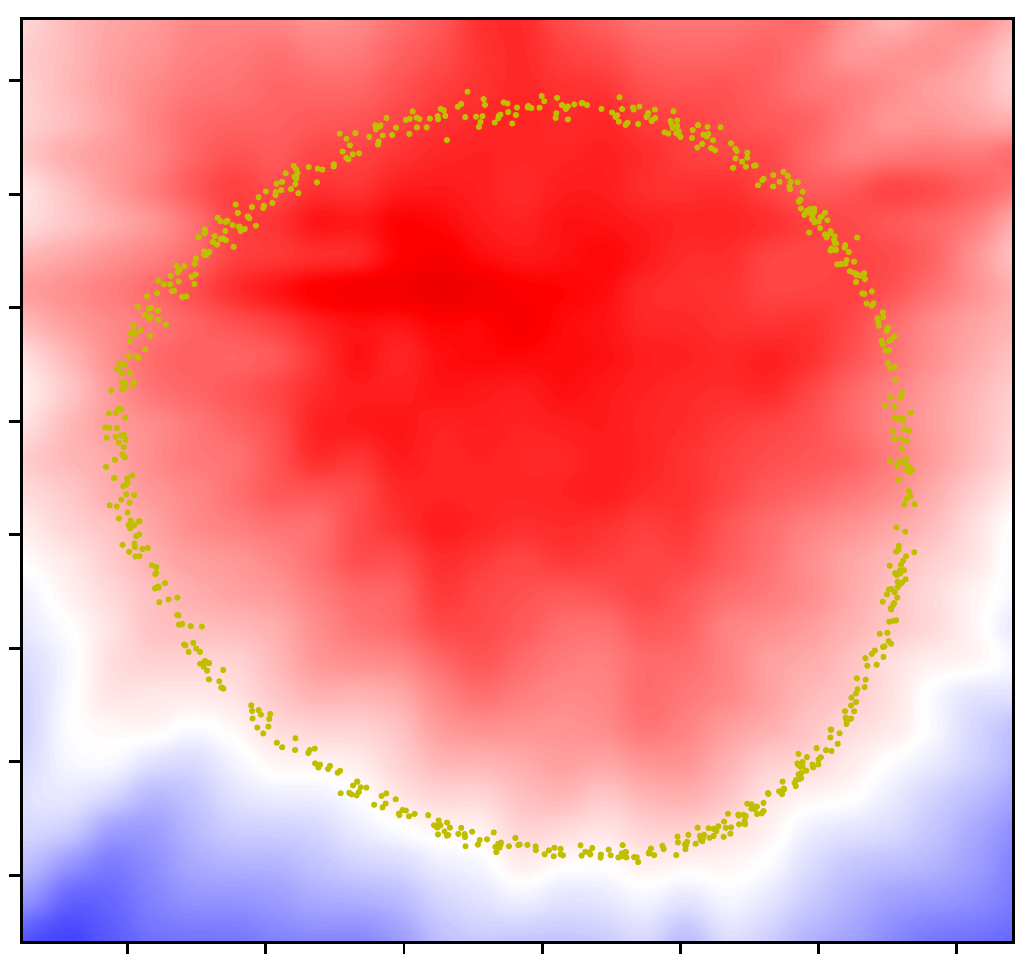}
		\includegraphics[width=0.23\textwidth]{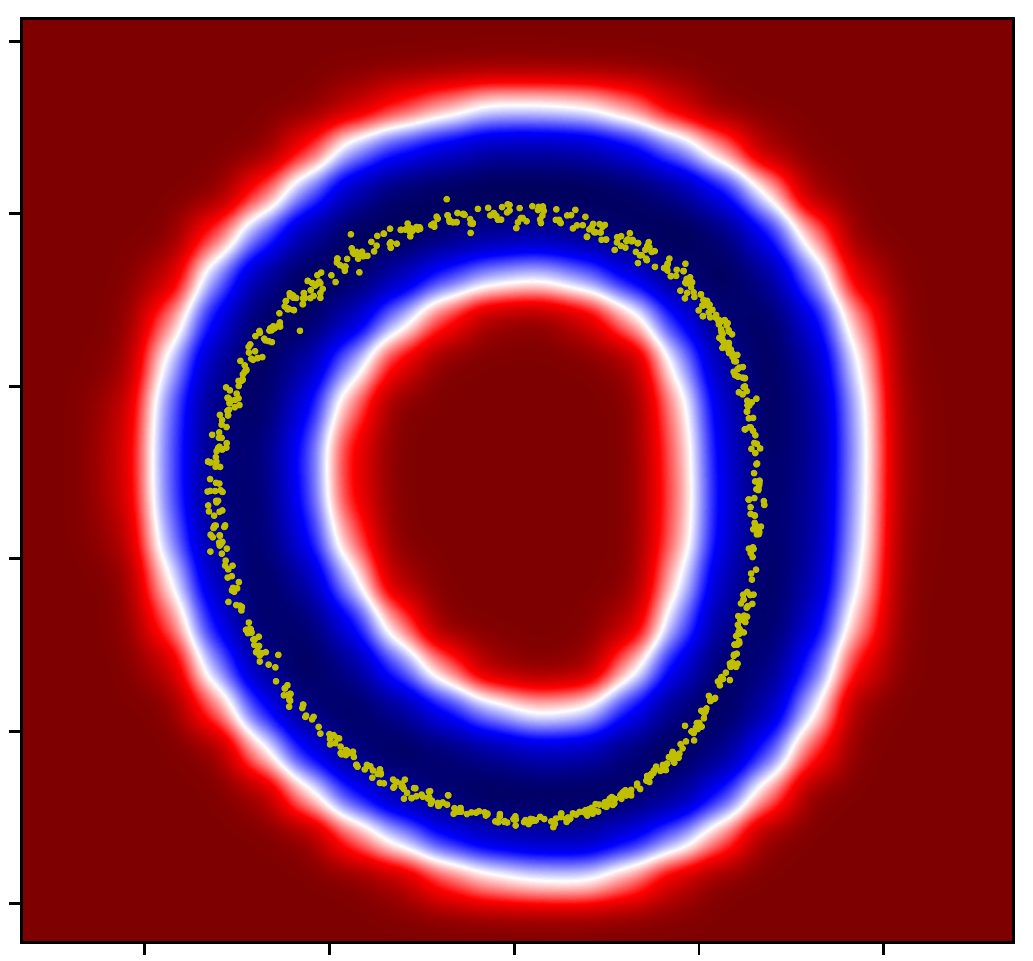}
		\includegraphics[width=0.23\textwidth]{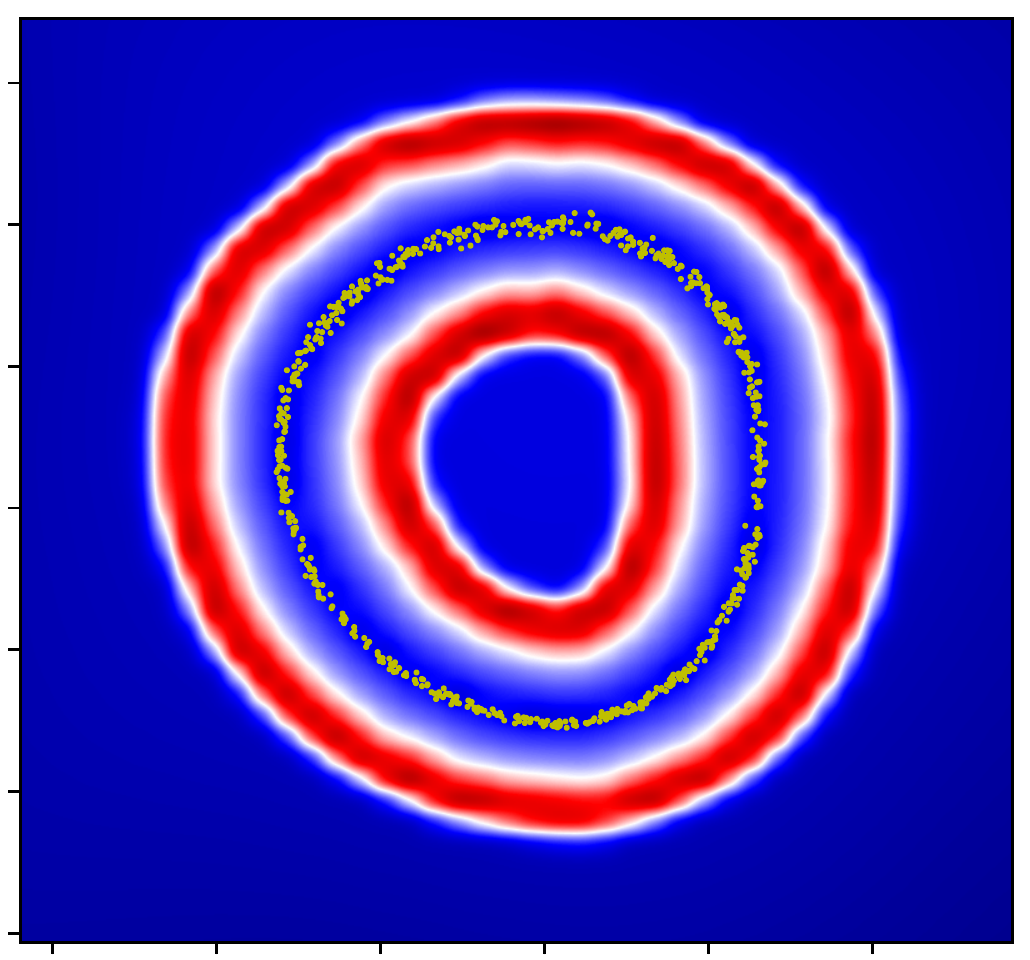}
		\caption{Inferred latent space for a toy data set, embedded via a non-linear function in $\mathbb{R}^{100}$. The background color, with blue representing lower and red representing higher values, from left to right, show: the (log) standard deviation estimated by a typical neural network; the associated (log) volume measure; the RBF (log) standard deviation estimate; and the associated (log) volume measure.
			Best viewed in color.
		}
		\label{fig:variance_plots}
	\end{figure*}
	
	\section{Geometric latent priors}\label{sec:priors}
	It is evident that the geometric structure over the latent space carries significant information about data density that the traditional Euclidean interpretation foregoes. With this in mind, we propose that the prior should be defined with respect to the geometric structure. We could opt for a \emph{Riemannian normal distribution}, which is well-studied \cite{oller1993intrinsic, mardia2000basic, pennec2006intrinsic, arvanitidis2016locally, hauberg2018directional}. Unfortunately, computing its normalization constant is expensive and involves Monte Carlo integration. Furthermore, it is equally hard to sample from this distribution, since it generally requires rejection sampling with non-trivial proposal distributions.
	
	Instead we consider a cheap and flexible alternative, namely the heat kernel of a Brownian motion process \cite{hsu2002stochastic}. A Brownian motion $X_t$ on an immersed Riemannian manifold $\mathcal{M} \subseteq \mathbb{R}^M$ can be defined through a stochastic differential equation on Stratonovich form: 
	\begin{align} 
	\mathrm{d}X_t = \sum_{\alpha=1}^M P_{\alpha}(X_t) \circ \mathrm{d}W_t^{\alpha}.
	\end{align}
	Here $W_t=(W_t^1,\dots,W_t^M)$ is a Brownian motion in $\mathbb{R}^M$ and $P_1(X_t),\dots,P_M(X_t)$ denotes the projection of the standard basis of $\mathbb{R}^M$ onto the tangent space of $\mathcal{M}$ at $X_t$. This way, a Brownian motion on $\mathcal{M}$ is driven by a Euclidean Brownian motion $W_t$ projected to the tangent space. Fixing an initial point $\boldsymbol{\mu} \in \mathcal{M}$ and a time $t>0$, Brownian motion starting at $\boldsymbol{\mu}$ running for time $t$ gives rise to a random variable on $\mathcal{M}$. Its density function is the \emph{transition density} $p(\x)$. An alternative description of Brownian motion on $\mathcal{M}$ is that $p(\x)$ is the heat kernel associated to the Laplace-Beltrami operator of a scalar function $h$ on $\mathcal{M}$:
	\begin{align}
	\Delta h=\dif{\M}^{-1} \partial_{i}\left(\dif{\M} g^{i j} \partial_{j} h\right)
	\end{align} 
	
	where $\dif{\M}$ is the volume measure of the immersed submanifold $\mathcal{M}$, $g^{i j}$ are the components of the inverse metric tensor and $\partial_{i}:=\frac{\partial}{\partial x^{i}}, \partial_{j}:=\frac{\partial}{\partial x^{j}}$ are the basis vectors at the tangent space $T_p\M$.
	We will express the transition density in terms of local coordinates $\mathcal{Z} \to \mathcal{M}$ on $\mathcal{M}$. Conveniently, we may approximate the transition density by a so-called Parametrix expansion in a power series \citep{hsu2002stochastic}. In this paper we will use the zeroth order approximation which gives rise to the following expression for $p(\z)$ with $\z \in \mathcal{Z}$:
	\begin{align}
	p(\z) & \approx
	(2\pi t)^{-\sfrac{d}{2}} H_0 \exp\left( -\frac{l^2(\z, \boldsymbol{\mu})}{2t} \right), \label{BMdensity}
	\end{align}
	where:
	\begin{itemize}
		\item $t \in \mathbb{R}$, denotes the duration of the Brownian motion, and corresponds to \emph{variance} on Euclidean manifolds.
		\item $d$ is the dimensionality of $\z$.
		\item $\boldsymbol{\mu} \in \mathcal{Z}$ is the center of the Brownian motion.
		\item $l(\cdot,\cdot)$ is the geodesic distance on the manifold.
		\item $H_0 = (\frac{\det \mat{G}_{\vec{z}}}{\det \mat{G}_{\boldsymbol{\mu}}})^{1/2}$ is the ratio of the Riemannian volume measure evaluated at points $\z$ and $\boldsymbol{\mu}$ respectively.
	\end{itemize}
	Equation~\ref{BMdensity} can be evaluated reasonably fast as no Monte Carlo integration is required. The most expensive computation is the evaluation of the geodesic distance for which several efficient algorithms exist \citep{hennig:aistats:2014, arvanitidis:aistats:2019}. Here we parameterize the geodesic as a cubic spline and perform direct energy minimization.
	
	\subsection{Inference}
	Since we use the heat kernel density function for the prior $p(\z)$, we need the variational family $q_\phi(\z|\x)$ to be defined with respect to the same Riemannian measure. We therefore also use the heat kernel density function for the variational family, which is parameterized by the encoder network with variational parameters $\phi$. The parameter $t$ of the prior is learned through optimization. The ELBO can be derived with respect to the volume measure $\dif \mathcal{M}$: 
	\begin{align}
	\log &p(\vec{x})  \geq \mathcal{L}_{\mathcal{M}}(\vec{x} ; \theta, \phi) \nonumber \\
	& \triangleq \int_{\mathcal{M}} \log  \left(\frac{p_{\theta}(\vec{x} | \vec{z}) p(\vec{z})}{q_{\phi}(\vec{z} | \vec{x})}\right) q_{\phi}(\vec{z} | \vec{x}) d \mathcal{M}_{\vec{z}} \nonumber \\
	&= \E_{q(\vec{z}|\vec{x})}[\log p_\theta(\vec{x|z})] - \KL{q_\phi(\z|\x)}{p(\z)}.
	\end{align}
	This ELBO can be estimated using Monte Carlo samples from the variational posterior. With no analytical solution to the KL divergence we resort to Monte Carlo integration:
	\begin{align}
	\KL{q}{p} &= \int_{\mathcal{M}} \log \frac{{q_\phi(\z|\vec{x})}}{p(\z)} q_\phi(\z|\vec{x}) d \mathcal{M}_\vec{z} \nonumber \\
	&= \E_{q(\vec{z}|\vec{x})}[\log q(\vec{z}|\vec{x}) - \log p(\vec{z})] \nonumber \\
	&\approx \frac{1}{N} \sum_{i=1}^{N} (\log q(\vec{z}_{i}|\vec{x}) - \log p(\vec{z}_{i}))
	\end{align}
	with:
	\begin{align}
	\log{q_\phi(\z|\vec{x})} &= -\frac{d}{2} \log (2\pi t_q) + \log{H_{0, q}} -\frac{l_q^2}{2t_q} \\ 
	\log{p(\z)} &= -\frac{d}{2} \log (2\pi t_p) + \log{H_{0, p}} -\frac{l_p^2}{2t_p}
	\end{align}
	where $l_q^2 = l^2(\z, \boldsymbol{\mu}_{q})$, $l_p^2 = l^2(\z, \boldsymbol{\mu}_{p})$.
	
	Thus, the final form of the Monte Carlo evaluation of the KL divergence is:
	\begin{align}
	&\KL{q}{p} \approx \frac{1}{2} \bigg[ \frac{1}{N} \sum_{i = 1}^{N} \bigg( \log \det \mat{G}_{\mu_p}(\vec{z}_i) - \nonumber \\
	&\log \det \mat{G}_{\mu_q}(\vec{z}_i) + \frac{l^2(\z_{i}, {\boldsymbol{\mu}_p})}{t_{p}} - \frac{l^2(\z_{i}, \boldsymbol{\mu}_{q})}{t_{q}} \bigg) \nonumber \\
	&+ d(\log t_p - \log t_q) \bigg]
	\end{align}
	
	\begin{figure}[h]
		\centering
		\includegraphics[width=0.46\textwidth]{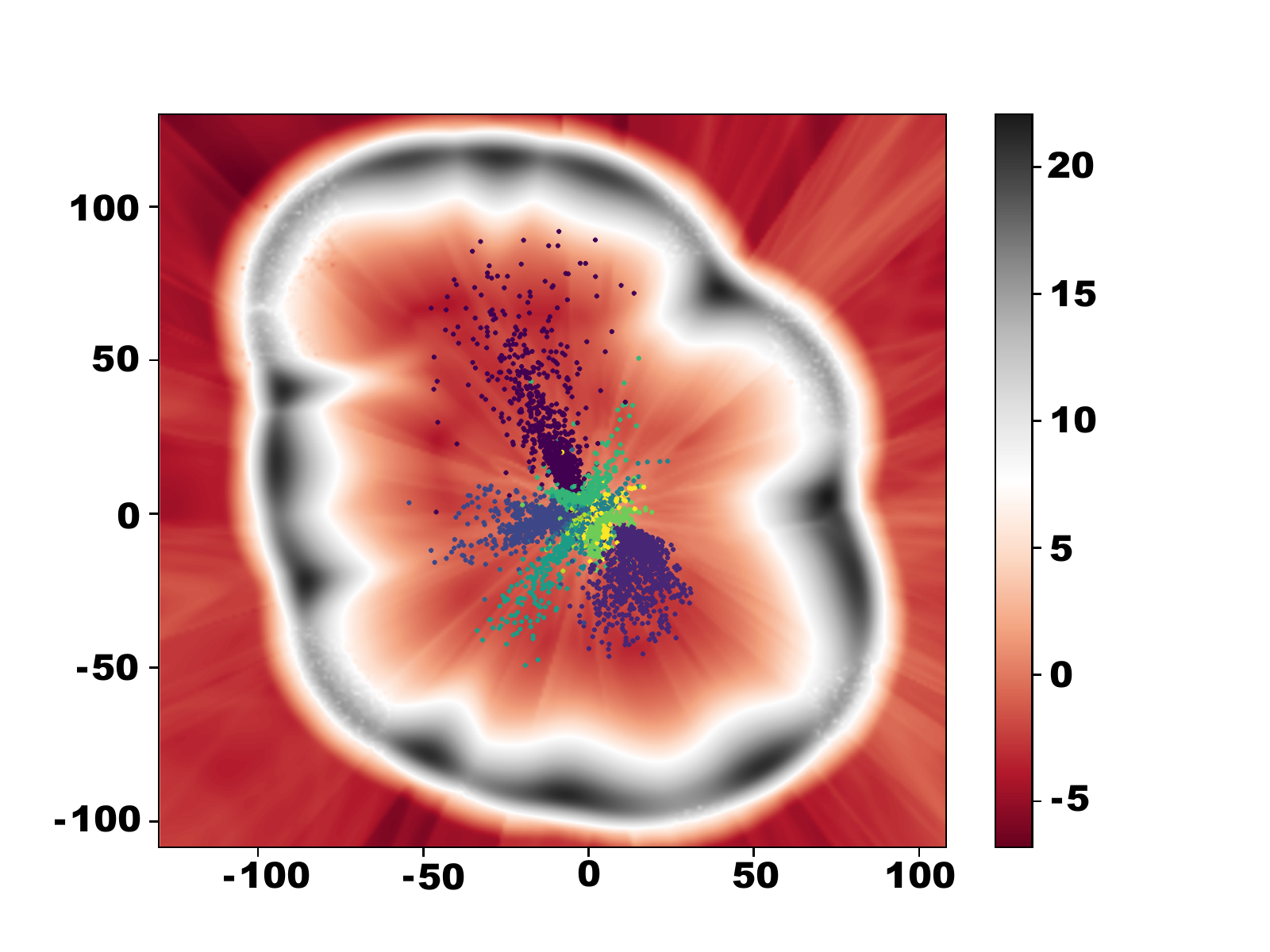}
		\caption{Latent space of an $\mathcal{R}$-VAE, plotted against the Riemannian volume measure $d\mathcal{M}$. Once again note the ``borders'' created by the metric roughly demarcating the latent code support. The latent codes are colored according to label. Best viewed in color.}
		\label{fig:RVAE_latent_space}
	\end{figure}
	\subsection{Sampling}
	In the previous section we mentioned that a Brownian motion (BM) on the manifold can be derived by projecting each BM step $X_t$ on the tangent space at $t$. However we will take each step directly in the latent space and avoid having to evaluate the exponential map. Given a manifold $\mathcal{M}$ with dimension $N$, the immersion $f: \mathcal{M} \rightarrow \mathbb{R}^{M}$, a point $\mathbf{a} \in \mathcal{M}$ and its image under $f$, $\boldsymbol{A} \in \mathbb{R}^{M}$ we take a random step from  $\boldsymbol{A}$:
	\begin{align}
	\Delta \sim \mathcal{N}\left(\mathbf{0}, \Sigma_{M}\right).
	\end{align}
	Applying a Taylor expansion we have:
	\begin{align}
	f(\mathbf{a}+\epsilon)=f(\mathbf{a})+\mathbf{J}_{\mathbf{a}} \epsilon+\mathcal{O}\left(\epsilon^{2}\right).
	\end{align}
	With $\Delta=f(\mathbf{a}+\epsilon)-f(\mathbf{a})$ we have:
	\begin{align}
	\Delta=\mathbf{J}_{\mathbf{a}} \epsilon+\mathcal{O}\left(\epsilon^{2}\right).
	\end{align}
	For small $\epsilon$ an approximation to taking a step directly in the latent space is then $\mathbf{b}=\mathbf{a}+\epsilon$ with $\epsilon \approx \mathbf{J}_{\mathrm{a}}^{+} \Delta$ and $\mathbf{J}_{\mathbf{a}}^{+}=\left(\mathbf{J}_{\mathbf{a}}^{\top} \mathbf{J}_{\mathbf{a}}\right)^{-1} \mathbf{J}_{\mathbf{a}}^{\top} \in \mathbb{R}^{N \times M}$ the pseudoinverse of $\mathbf{J}_{\mathbf{a}}$. Since $\Delta \sim \mathcal{N}\left(\mathbf{0}, \Sigma_{M}\right)$ the step $\epsilon$ can be written:
	\begin{align}
	\epsilon \sim \mathcal{N}\left(\mathbf{0}, \mathbf{J}_{\mathbf{a}}^{+} \Sigma_{M}\left(\mathbf{J}_{\mathbf{a}}^{+}\right)^{\top}\right).
	\end{align}
	We consider an isotropic heat kernel so in our case $\Sigma_{M}=\sigma^{2} \mathbf{I}$. Furthermore:
	\begin{align}
	\mathbf{J}_{\mathbf{a}}^{+} \Sigma_{M}&\left(\mathbf{J}_{\mathbf{a}}^{+}\right)^{\top} 
	= \left(\mathbf{J}_{\mathbf{a}}^{\top} \mathbf{J}_{\mathbf{a}}\right)^{-1} \mathbf{J}_{\mathbf{a}}^{\top} \Sigma_{M} \mathbf{J}_{\mathbf{a}}\left(\mathbf{J}_{\mathbf{a}}^{\top} \mathbf{J}_{\mathbf{a}}\right)^{-\top} \nonumber \\
	&= \sigma^2\left(\mathbf{J}_{\mathbf{a}}^{\top} \mathbf{J}_{\mathbf{a}}\right)^{-1} \mathbf{J}_{\mathbf{a}}^{\top} \mathbf{J}_{\mathbf{a}}\left(\mathbf{J}_{\mathbf{a}}^{\top} \mathbf{J}_{\mathbf{a}}\right)^{-\top} \nonumber \\
	&= \sigma^2\left(\mathbf{J}_{\mathbf{a}}^{\top}  \mathbf{J}_{\mathbf{a}}\right)^{-\top}
	= \sigma^2\left(\mathbf{J}_{\mathbf{a}}^{\top} \mathbf{J}_{\mathbf{a}}\right)^{-1}.
	\end{align}
	This implies that
	\begin{align}
	\epsilon \sim \mathcal{N}\left(\mathbf{0}, \sigma^{2}\left(\mathbf{J}_{\mathbf{a}}^{\top} \mathbf{J}_{\mathbf{a}}\right)^{-1}\right).
	\end{align}
	Thus, to sample from the prior we simply need to run Brownian motion for $t = 1, \ldots, T$:
	\begin{align}
	\z_t &\sim \N\left(\z_{t-1}, \frac{\sigma^{2}}{T} \left(\J_{\z_{t-1}}\T\J_{\z_{t-1}}\right)^{-1} \right)
	\end{align}
	
	An obvious concern regarding the computational cost of sampling is the inverting of the metric tensor. While this is a valid concern for large latent dimensionalities, in practice and for the typical number of latent dimensions found in generative modelling literature the sampling cost is bearable, considering that the operation can be parallelized for $K$ samples. We further note that from a practical standpoint for small diffusion times the number of discretized steps can be small. The time complexity of the sampling operation is
	\begin{align}
	\mathcal{O}(KHM + KMN^2 + N^3)
	\end{align}
	where $K$ is the number of samples, $N$ is the latent space dimensionality, $M$ is the input space dimensionality and $H$ is the decoder hidden layer size.
	\begin{table}[t]
		\centering
		\caption{Results on MNIST (mean \& std deviation over 10 runs). \textit{Rec} denotes the negative conditional likelihood.}
		\resizebox{\columnwidth}{!}{
			\begin{tabular}{ccccc}
				Model & Neg. ELBO & Rec & KL\\
				\toprule
				VAE &  &  &  \\ \hline
				d = 2 & -1030.38$_{\pm5.34}$ & -1033.06$_{\pm5.48}$ & 2.68$_{\pm.14}$\\
				d = 5 & -1076.64$_{\pm4.48}$ & -1078.91$_{\pm4.44}$ & 2.27$_{\pm.04}$\\
				d = 10 &  -1110.79$_{\pm1.17}$ & -1113.01$_{\pm1.13}$ & 2.22$_{\pm.03}$\\ \hline
				VAE-VampPrior &  &  &  & \\ \hline
				d = 2 & -1045.03$_{\pm5.22}$ & -1047.34$_{\pm5.20}$ & 2.30$_{\pm.03}$\\
				d = 5 & -1109.74$_{\pm4.87}$ & -1111.63$_{\pm4.87}$ & 1.88$_{\pm.01}$\\
				d = 10 & -1116.58$_{\pm4.23}$ & -1118.27$_{\pm4.20}$ & 1.69$_{\pm.02}$\\ \hline
				$\mathcal{R}$-VAE &  &  &  & \\ \hline
				d = 2 & \textbf{-1047.29$_{\pm2.77}$} & \textbf{-1053.70$_{\pm2.75}$} & 14.33$_{\pm.01}$\\
				d = 5 & \textbf{-1141.06$_{\pm7.09}$} & \textbf{-1177.86$_{\pm3.39}$} & 28.00$_{\pm.2}$ \\
				d = 10 & \textbf{-1170.03$_{\pm18.52}$} & \textbf{-1280.94$_{\pm14.67}$} & 57.76$_{\pm3.85}$ \\ \bottomrule
		\end{tabular}}
		\label{tab:mnist}
	\end{table}
	
	\section{Meaningful variance estimation} \label{variance_estimation}
	We now turn to the problem of restricting our prior to sample from the image of our manifold in $\mathcal{Z}$. Since typically the geometry of the data is not known a priori, we adopt the Bayesian approach and relate uncertainty estimation in the generative model to the geometry of the latent manifold. Specifically, since the generative model parameterizes $f_\theta: \mathcal{Z} \rightarrow \mathcal{X}$ we construct it such that the pull-back metric will acquire high values away from the data support and thereby restrict prior samples to high density regions of the latent manifold.
	
	In Sec.~\ref{immersions} we described the metric tensor arising from the diagonal immersion $f$. By the form of the metric, it is clear that both $\mu_\theta(\z)$ and $\sigma_\theta(\z)$ contribute to the manifold geometry. In recent works \cite{arvanitidis:iclr:2018, hauberg2018only, detlefsen2019reliable} it was shown that neural network variance estimates are typically poor in regions away from the training data, due to poor extrapolation properties. Thus, neural networks cannot be trusted to properly estimate the variance of the generative model ``off-the-shelf'' when the functional form of the immersion (and thus the geometry of the data) is not known a priori. By extension, this leads to poor estimates of latent manifold geometry and latent densities. \citet{arvanitidis:iclr:2018} propose to use a \emph{radial basis function (RBF)} network \cite{que:aistats:2016} to estimate precision, rather than variance. We adopt this approach due to its simplicity and relative numerical stability, however we note that similar approaches for principled variance estimation exist \cite{detlefsen2019reliable, stirn2020variational}.
	
	The influence of the RBF network can be seen in Fig.~\ref{fig:variance_plots}, where it is compared with a usual neural network variance estimate. Note that the metric creates ``borders'' demarcating the regions to which the latent codes have been mapped by the encoder. This makes interpolations and random walks generally follow the trend of the latent points instead of wondering off the support. Thus, this regularization scheme restricts prior sampling to such high density regions. A similar effect is not observed in the usual Gaussian VAE, where the prior samples from regions to which the variational posterior has not necessarily placed probability density \cite{hoffman2016elbo, rosca2018distribution}.
	
	\begin{table}[t]
		\centering
		\caption{Results on FashionMNIST (mean \& std deviation over 10 runs). \textit{Rec} denotes the negative conditional likelihood.}
		\resizebox{\columnwidth}{!}{
			\begin{tabular}{ccccc}
				Model & Neg. ELBO & Rec & KL\\
				\toprule
				VAE &  &  &  \\ \hline
				d = 2 & -443.13$_{\pm10.67}$ & -447.44$_{\pm10.8}$ & 4.31$_{\pm.14}$\\
				d = 5 & -511.65$_{\pm3.70}$ & -517.41$_{\pm3.84}$ & 5.76$_{\pm.21}$\\
				d = 10 & -525.05$_{\pm5.87}$ & -530.86$_{\pm5.9}$ & 5.81$_{\pm.05}$\\ \hline
				VAE-VampPrior &  &  &  & \\ \hline
				d = 2 & -705.90$_{\pm17.3}$ & -708.45$_{\pm17.29}$ & 2.54$_{\pm.01}$\\
				d = 5 & -769.27$_{\pm5.}$ & -770.1$_{\pm5.02}$ & 0.83$_{\pm.09}$\\
				d = 10 & -774.17$_{\pm10.83}$ & -777.75 $_{\pm10.78}$ & 3.57$_{\pm.06}$\\ \hline
				$\mathcal{R}$-VAE &  &  &  & \\ \hline
				d = 2 & \textbf{-708.77$_{\pm6.93}$} & \textbf{-722.41$_{\pm5.736}$} & 13.64$_{\pm1.51}$\\
				d = 5 & \textbf{-889.62$_{\pm3.44}$} & \textbf{-913.61$_{\pm3.38}$} & 23.83$_{\pm.8}$ \\
				d = 10 & \textbf{-959.2$_{\pm5.37}$} & \textbf{-1001.4$_{\pm4.08}$} & 40.35$_{\pm.8}$ \\ \bottomrule
		\end{tabular}}
		\label{tab:fmnist}
	\end{table}
	
	\section{Experiments}
	\begin{table*}[t]
		\caption{Per digit and average F1 score for a classifier trained on the learned latent codes of VAE and $\mathcal{R}$-VAE. Results are averaged over 5 classifier training runs.}
		\centering
		\begin{tabular}{cccccccccccc}
			Digits & 0 & 1 & 2 & 3 & 4 & 5 & 6 & 7 & 8 & 9 & Avg \\
			\toprule
			VAE & & & & & & & & & & & \\ \hline
			d = 2 & 0.94 & 0.95 & 0.88 & 0.67 & 0.55 & 0.42 & 0.86 & 0.68 & 0.61 & 0.53 & 0.72$_{\pm.002}$ \\
			d = 5 & 0.95 & 0.97 & 0.94 & 0.90 & 0.90 & 0.89 & 0.95 & 0.93 & 0.88 & 0.87 & 0.92$_{\pm.001}$ \\
			d = 10 & 0.98 & 0.99 & 0.97 & 0.94 & 0.96 & 0.95 & 0.98 & 0.97 & 0.93 & 0.94 & \textbf{0.96$_{\pm.001}$} \\ \hline
			$\mathcal{R}$-VAE & & & & & & & & & & & \\ \hline
			d = 2 & 0.95 & 0.97 & 0.89 & 0.68 & 0.64 & 0.56 & 0.88 & 0.85 & 0.71 & 0.64 & \textbf{0.78$_{\pm.002}$} \\
			d = 5 & 0.95 & 0.98 & 0.94 & 0.91 & 0.94 & 0.88 & 0.95 & 0.93 & 0.90 & 0.89 & \textbf{0.93$_{\pm.0008}$} \\
			d = 10 & 0.98 & 0.98 & 0.96 & 0.95 & 0.96 & 0.95 & 0.97 & 0.97 & 0.93 & 0.94 & \textbf{0.96$_{\pm.001}$} \\
		\end{tabular}
		\label{tab:classification}
	\end{table*}
	
	\begin{figure*}[h!]
		\centering
		\begin{subfigure}[t]{0.48\textwidth}
			\centering
			\includegraphics[width=1.0\textwidth]{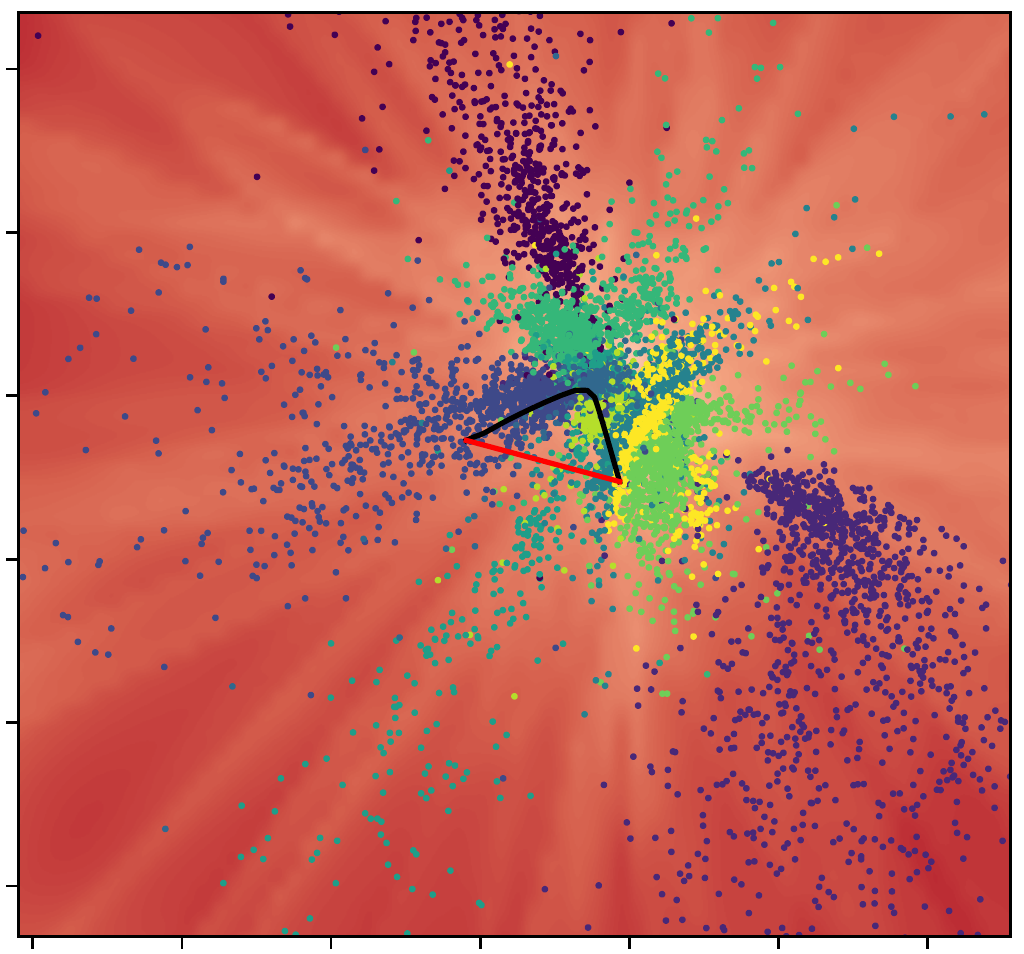}
		\end{subfigure}
		\begin{subfigure}[t]{0.48\textwidth}
			\centering
			\includegraphics[width=1.0\textwidth]{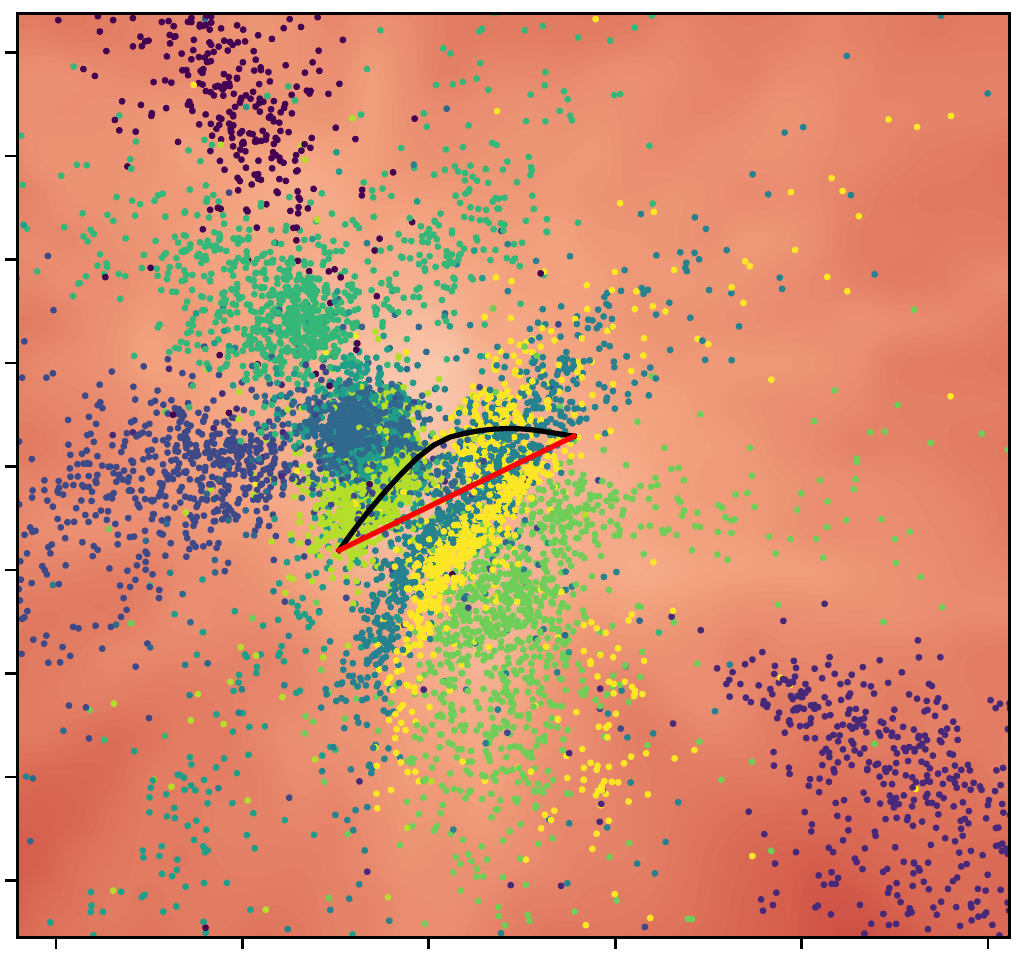}
		\end{subfigure}
		
		\begin{subfigure}[t]{0.48\textwidth}
			\centering
			\includegraphics[width=1.0\textwidth]{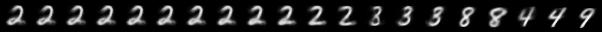}
		\end{subfigure}
		\begin{subfigure}[t]{0.48\textwidth}
			\centering
			\includegraphics[width=1.0\textwidth]{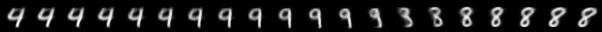}
		\end{subfigure}
		
		\begin{subfigure}[t]{0.48\textwidth}
			\centering
			\includegraphics[width=1.0\textwidth]{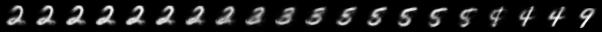}
		\end{subfigure}
		\begin{subfigure}[t]{0.48\textwidth}
			\centering
			\includegraphics[width=1.0\textwidth]{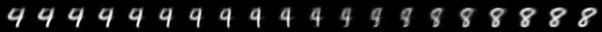}
		\end{subfigure}
		\caption{\textit{Top}: Interpolations plotted in the latent space of $\mathcal{R}$-VAE. Black indicates a geodesic interpolant, red indicates a Euclidean interpolant. \textit{Middle}: Images reconstructed along the geodesic interpolation. \textit{Bottom}: Images reconstructed along the Euclidean interpolation. The latent codes are color-coded according to label. Best viewed in color.}
		\label{fig:interpolations}
	\end{figure*}
	
	\subsection{Generative modelling}
	For our first experiment we train a VAE with a Riemannian Brownian motion prior ($\mathcal{R}$-VAE) for different latent dimensions and compare it to a VAE with a standard Normal prior and a VAE with a VampPrior. Tables~\ref{tab:mnist} \& \ref{tab:fmnist} show the results. $\mathcal{R}$-VAE achieves a better lower bound than both its Euclidean counterparts. The Brownian motion prior adapts to the latent code support and as such yields more expressive representations. On the other hand, with only a single parameter it results in a model that generalizes better than VAEs with a VampPrior.
	
	\subsection{Classification}
	We next assess the usefulness of the latent representations of $\mathcal{R}$-VAE. Fig.~\ref{fig:RVAE_latent_space} shows the latent code clusters. $\mathcal{R}$-VAE has produced more separable clusters in the latent space due to the prior adapting to the latent codes, which results in a less regularized clustering. We quantitatively measured the utility of the $\mathcal{R}$-VAE latent codes in different dimensionalities by training a classifier to predict digit labels and measuring the average overall and per-digit F1 score. Table~\ref{tab:classification} shows the results when comparing against the same classifier trained on latent codes derived by a VAE. $\mathcal{R}$-VAE has a significant advantage in low dimensions. As dimensionality increases this advantage becomes non-existent. An explanation for this is that due to the KL annealing of the Euclidean VAE, its representations have become more informative.
	
	\subsection{Qualitative results}
	\begin{figure*}[h!]
		\centering
		\begin{subfigure}[t]{0.3\textwidth}
			\centering
			\includegraphics[width=.93\textwidth]{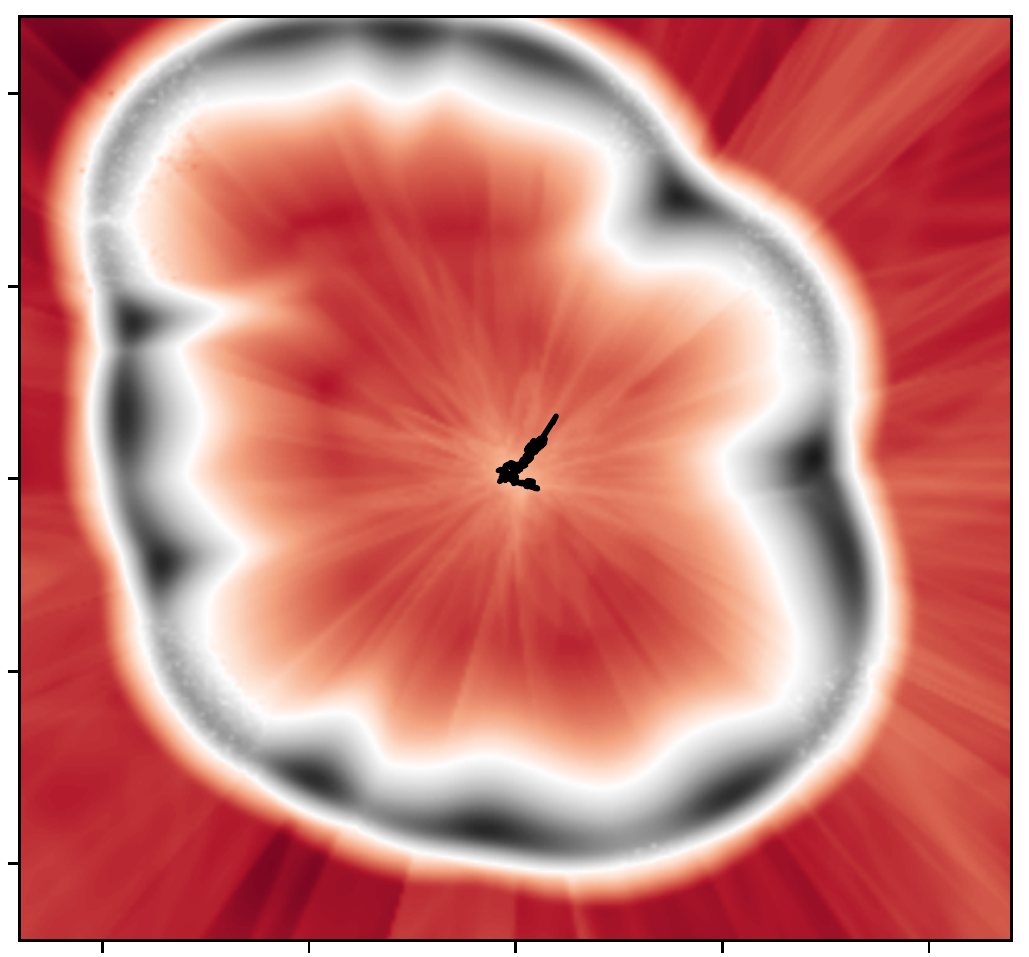}
		\end{subfigure}
		\begin{subfigure}[t]{0.3\textwidth}
			\centering
			\includegraphics[width=.93\textwidth]{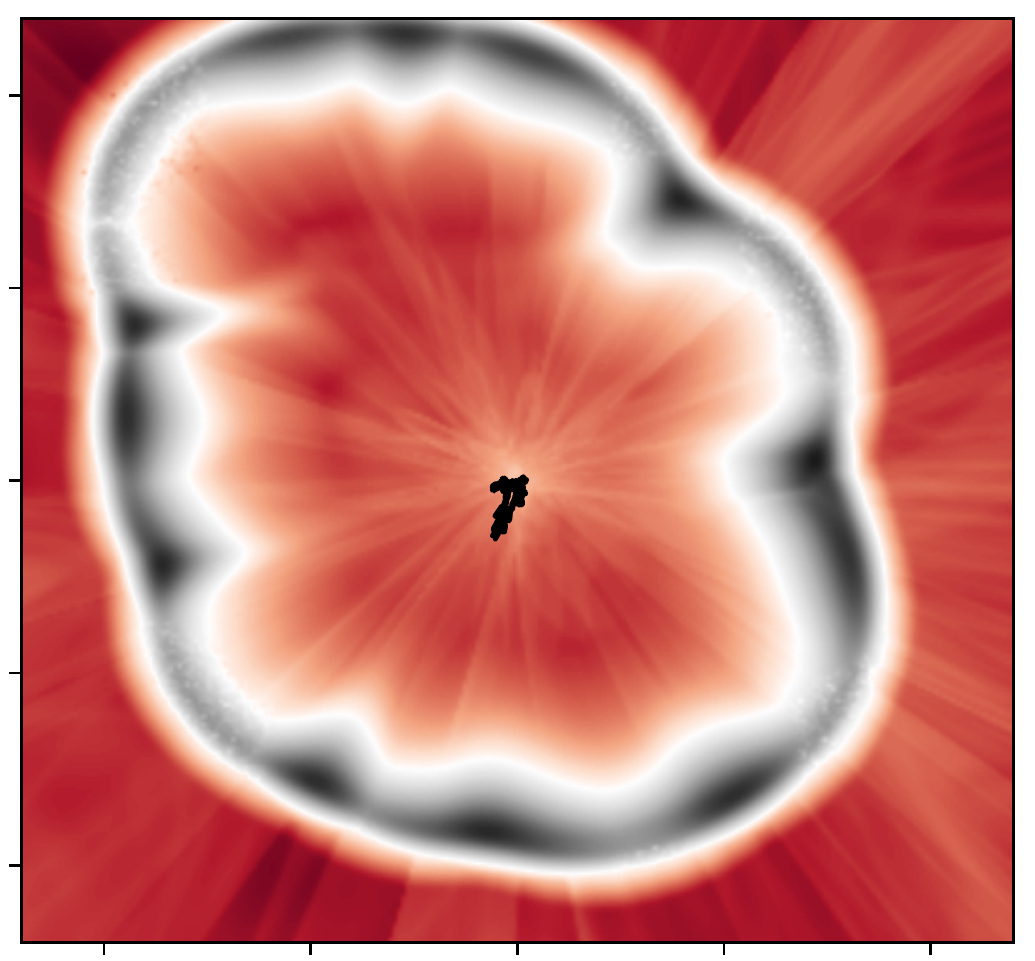}
		\end{subfigure}
		\begin{subfigure}[t]{0.3\textwidth}
			\centering
			\includegraphics[width=.93\textwidth]{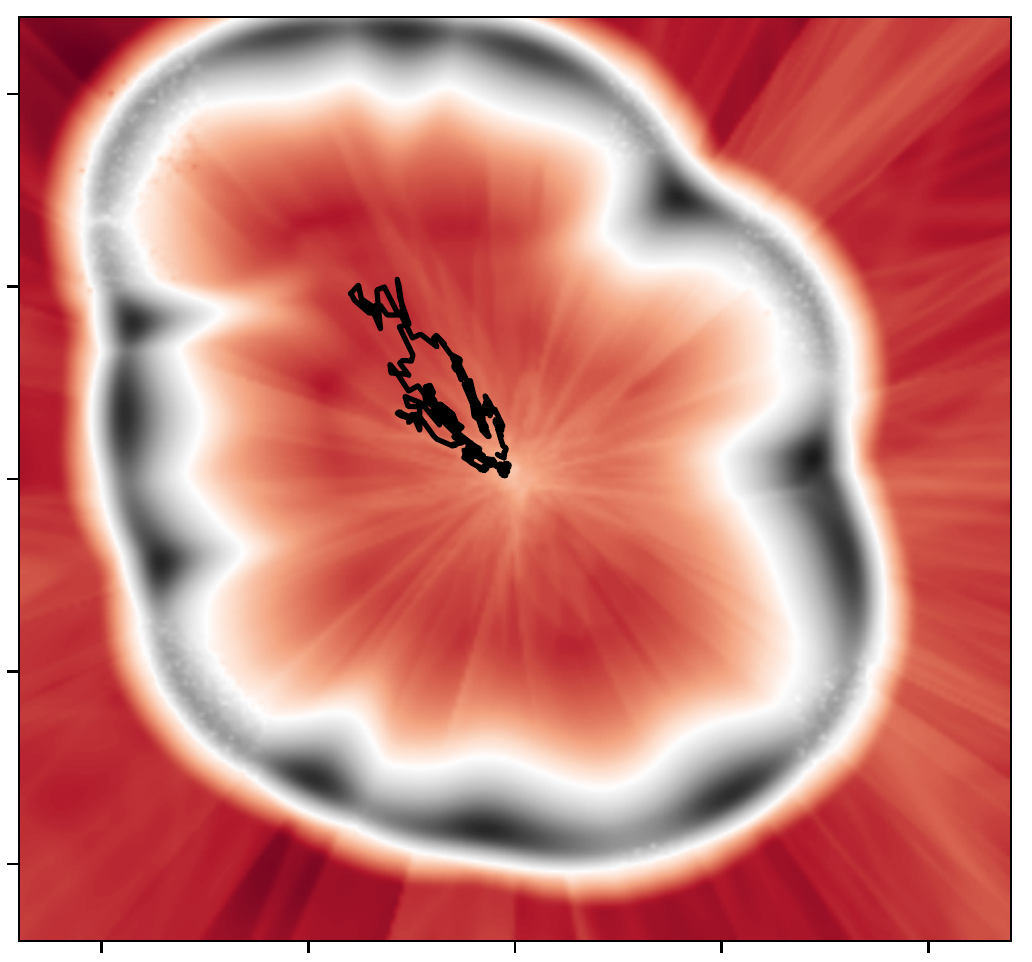}
		\end{subfigure}
		
		\begin{subfigure}[t]{0.3\textwidth}
			\centering
			\includegraphics[width=.9\textwidth]{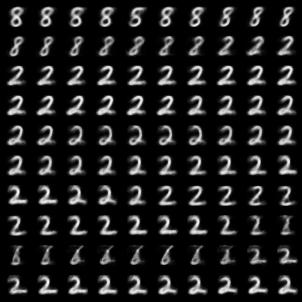}
		\end{subfigure}
		\begin{subfigure}[t]{0.3\textwidth}
			\centering
			\includegraphics[width=.9\textwidth]{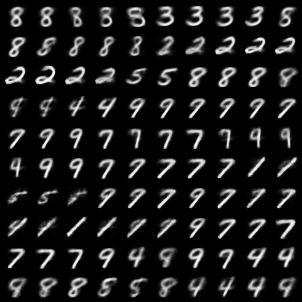}
		\end{subfigure}
		\begin{subfigure}[t]{0.3\textwidth}
			\centering
			\includegraphics[width=.9\textwidth]{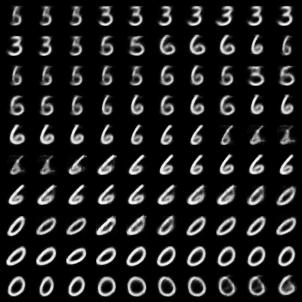}
		\end{subfigure}
		\caption{\textit{Top}: Brownian motion runs on the learned latent manifold. \textit{Bottom}: Corresponding sampled images. The sampler mostly stays in high density regions of the latent manifold. Best viewed in color.}
		\label{fig:samples}
	\end{figure*}
	\begin{figure*}[t!]
		\centering
		\begin{subfigure}[t]{0.3\textwidth}
			\centering
			\includegraphics[width=.93\textwidth]{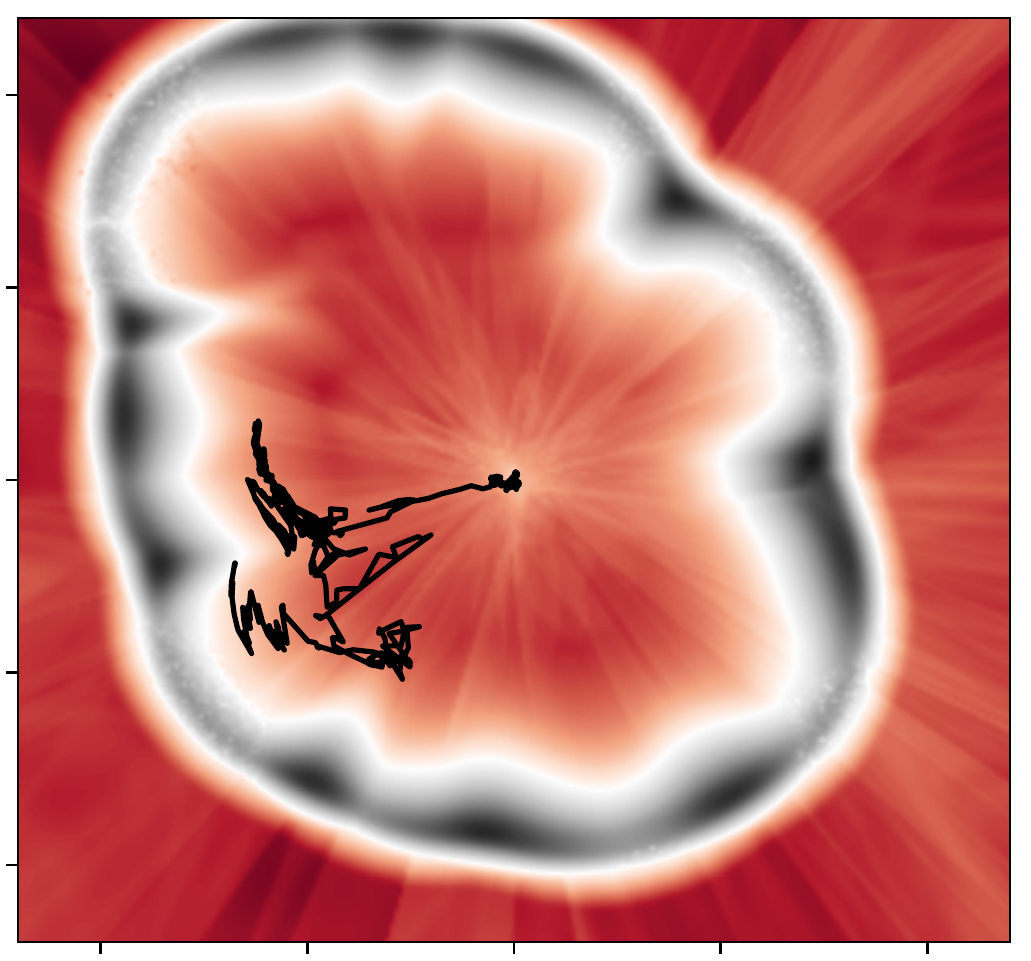}
		\end{subfigure}
		\begin{subfigure}[t]{0.3\textwidth}
			\centering
			\includegraphics[width=.93\textwidth]{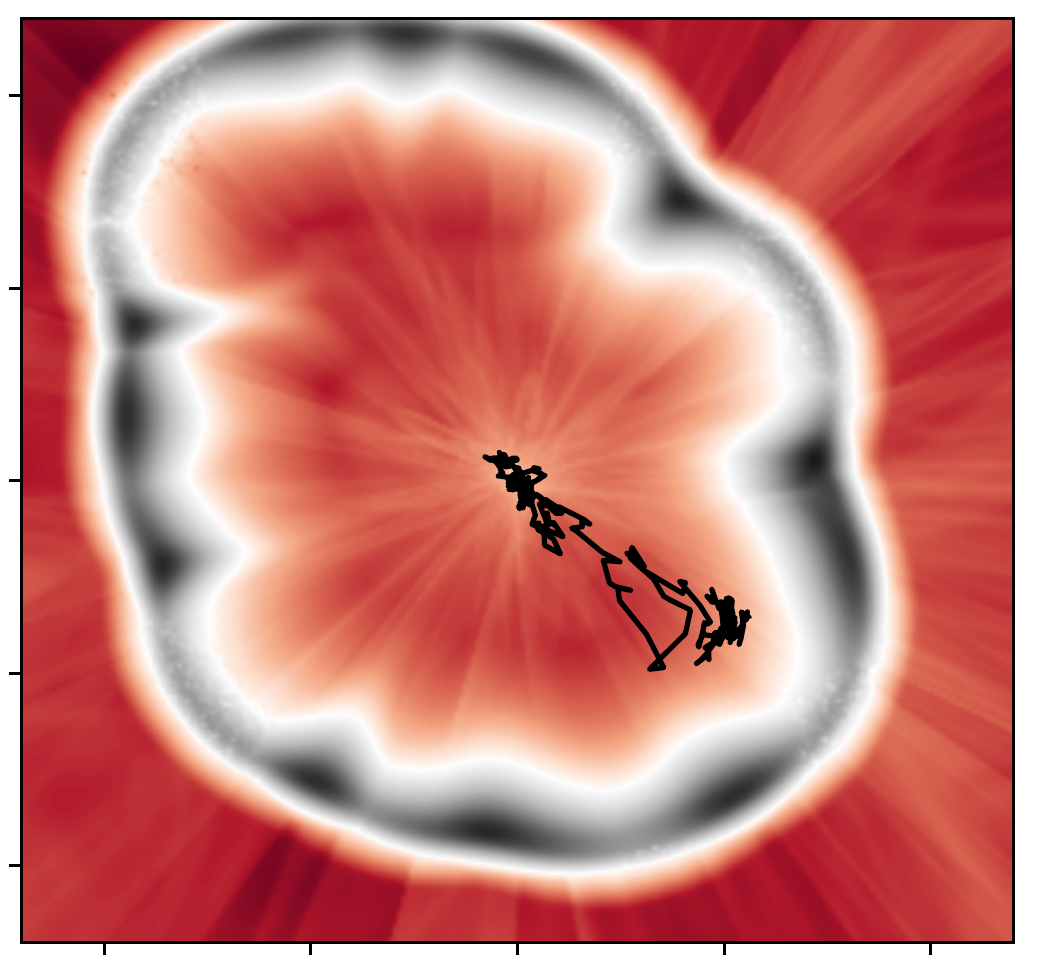}
		\end{subfigure}
		\begin{subfigure}[t]{0.3\textwidth}
			\centering
			\includegraphics[width=.93\textwidth]{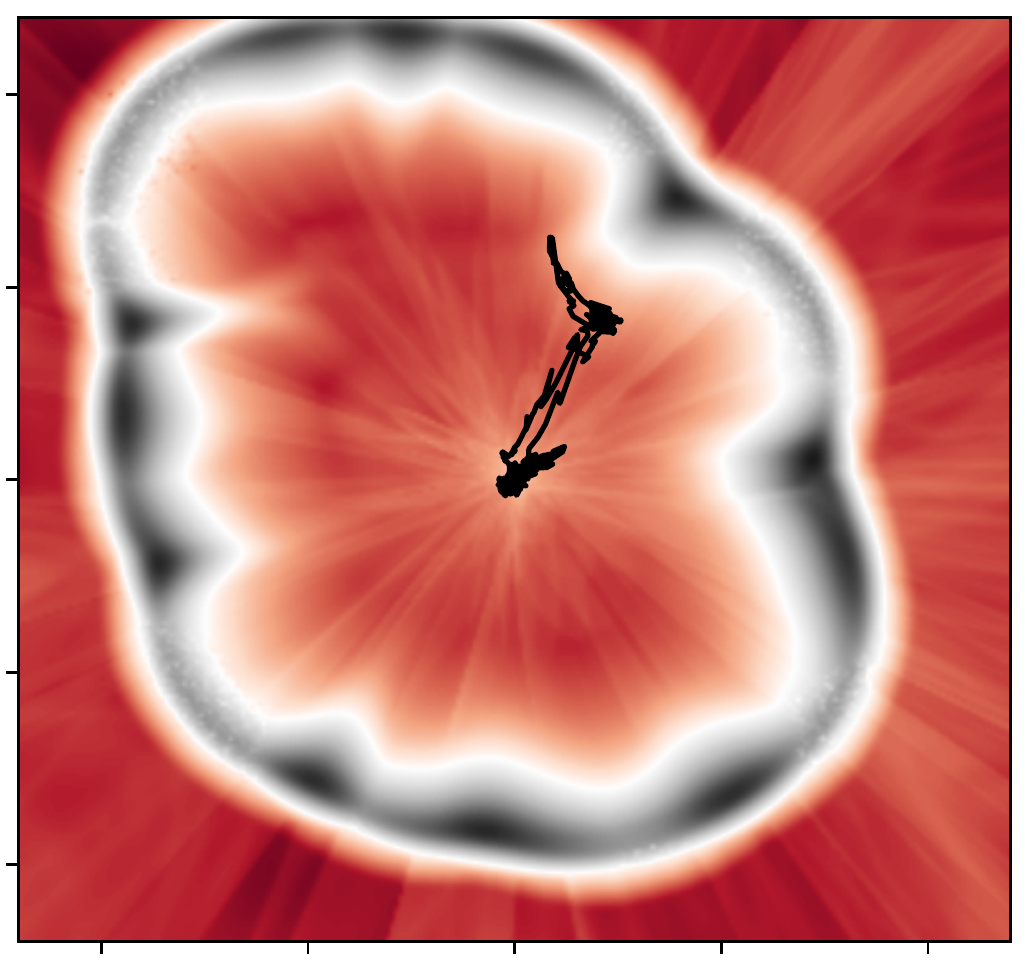}
		\end{subfigure}
		
		\begin{subfigure}[t]{0.3\textwidth}
			\centering
			\includegraphics[width=.9\textwidth]{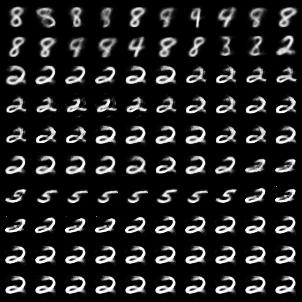}
			\caption{}
			\label{fig:fail_a}
		\end{subfigure}
		\begin{subfigure}[t]{0.3\textwidth}
			\centering
			\includegraphics[width=.9\textwidth]{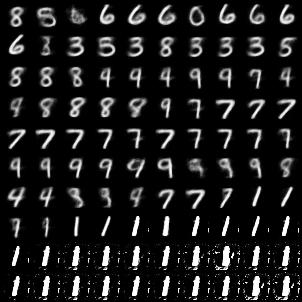}
			\caption{}
			\label{fig:fail_b}
		\end{subfigure}
		\begin{subfigure}[t]{0.3\textwidth}
			\centering
			\includegraphics[width=.9\textwidth]{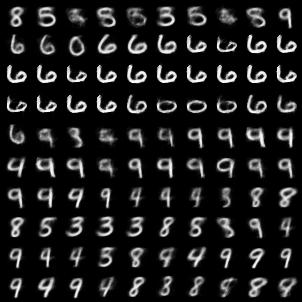}
			\caption{}
			\label{fig:fail_c}
		\end{subfigure}
		\caption{Brownian motion runs with artificially increased $t$ (diffusion) parameter beyond the learned value. Note that the borders created by the metric tensor stop the sampler from exploring low density regions any further - the sampler either stops (a and b) or returns to regions of higher density (c). This effect is observed in the sampled images. Best viewed in color.}
		\label{fig:fail_samples}
	\end{figure*}
	Finally we explore the geometric properties of a $\mathcal{R}$-VAE with a 2-dimensional latent space. Fig~\ref{fig:RVAE_latent_space} shows the learned manifold. As in Fig.~\ref{fig:variance_plots}, the influence of the variance network on the metric can be seen in the ``borders'' surrounding the latent code support.
	
	We begin by investigating the behavior of distances on the induced manifold. Fig.~\ref{fig:interpolations} shows the geodesic curves between two pairs of random points on the manifold, compared against their Euclidean counterpart. The geodesic interpolation is influenced by the metric tensor, which makes sure that shortest paths will generally avoid areas of low density. This can easily be seen in top left Fig.~\ref{fig:interpolations}, where the geodesic curve follows a path along a high density region. Contrast this to the Euclidean straight line between the two points traversing a lower density region. Reconstructed images along the curves can be seen in the middle and bottom rows. Even in less apparent cases (top right Fig.~\ref{fig:interpolations}), reconstructions of latent codes along geodesic curves generally provide smoother transitions between the curve endpoints as can be seen by comparing the middle right and bottom right sections of the figure. 
	
	Next, we investigate sampling from $\mathcal{R}$-VAE. In Sec.~\ref{variance_estimation} we claimed that a Brownian motion prior coupled with the RBF regularization of the decoder variance network would yield samples that mostly avoid low density regions of the latent space. To empirically prove this, we executed two sets of multiple sampling runs on the latent manifold. In the first set we ran Brownian motion with the learned prior parameters. These runs and the resulting images are displayed in Fig.~\ref{fig:samples}. The random walks generally stay within high density regions of the manifold. Cases where they explore low density regions do exist but they are rare. The samples generally seem clear although sometimes their quality drops, especially when the sampler is transitioning between classes, where variance estimates are higher. This could potentially be rectified with a less aggressive deterministic warm-up scheme, which would result in more concentrated densities with thinner tails, although between-class variance estimates would likely still be higher compared to within-class ones. For the second set of the sampling runs, we increased the duration of the Brownian motion. These runs are displayed along with the sampled images in Fig.~\ref{fig:fail_samples}. The influence of the variance estimates on the metric tensor is clearly shown here. As the sampler is moving farther away from the latent code support, evaluations of the metric tensor increase making these regions harder to traverse. As a result the random walk either oscillates with decreased speed and stops close to the boundary (as in Figures~\ref{fig:fail_a} and \ref{fig:fail_b}) or returns to higher density regions of the manifold. This clearly shows that $\mathcal{R}$-VAE mostly avoids the manifold mismatch problem.
	
	\section{Related work}
	\paragraph{Learned priors.} In recent literature many works have identified the adverse effects of the KL divergence regularization when the prior is chosen to be a standard Gaussian. As such, there have been many approaches of learning a more flexible prior. \citet{chen2016variational} propose learning an autoregressive prior by applying an Inverse Autoregressive transformation \cite{kingma2016improved} to a simple prior. \citet{nalisnick2016stick} propose a non-parametric stick-breaking prior. \cite{tomczak2017vae} propose learning the prior as a mixture of variational posteriors. More recently, \citet{bauer2018resampled} present a rejection sampling approach with a learned acceptance function, while \citet{klushyn2019learning} proposed a hierarchical prior through an alternative formulation of the objective.
	
	\paragraph{Non-Euclidean latent space.} \citet{arvanitidis:iclr:2018} was one of the first to analyze the latent space of a VAE from a non-Euclidean perspective. This work was inspired by \citet{Tosi:UAI:2014} that studied the Riemannian geometry of the Gaussian process latent variable model \citep{gplvm}. \citet{arvanitidis:iclr:2018} train a Euclidean VAE and fit a latent Riemannian LAND distribution \cite{arvanitidis2016locally} and show that this view of the latent space leads to more accurate statistical estimates, as well as better sample quality.
	
	Since then, a number of other works have appeared in literature that propose learning non-Euclidean latent manifolds. \citet{xu2018spherical} and \citet{davidson2018hyperspherical} learn a VAE with a von Mises-Fisher latent distribution, which samples codes on the unit hypersphere. Similarly, \citet{mathieu2019continuous} and \citet{nagano2019differentiable} extend VAEs to hyperbolic spaces. \citet{mathieu2019continuous} assume a Poincar\'{e} ball model as a latent space and present 2 generalizations of the Euclidean Gaussian distribution - a wrapped Normal and the Riemannian Normal distributions, of which only the latter is a maximum entropy generalization. In practice, they perform similarly. \citet{nagano2019differentiable} assume a Lorentz hyperbolic model as a latent space and also present a wrapped Normal generalization of the Gaussian. While these works have correctly identified the problem of the standard Gaussian not being a truly uninformative prior, due to the Euclidean assumption, they have proposed approaches which are designed for observations with known geometries. Most of the time, however, this information is not available and a more general framework for learning geometrically informed VAEs is needed. In response to this, \citet{skopek2019mixed} propose VAEs with the latent space modelled as a product of constant cuvature manifolds, where each component curvature is learned. While more general than a model with a fixed curvature latent manifold, this framework still requires the specification of number of component manifolds along with the sign of their respective curvature. Finally, similar to our approach,  \citet{li2019variational} and \citet{rey2019diffusion} both propose the heat kernel as a variational family representing a Brownian motion process on a Riemannian manifold. They test their approaches on a priori chosen manifolds.
	
	\section{Conclusion}
	In this paper we presented VAEs with Riemannian manifolds as latent spaces and proposed a Riemannian generalization of the Gaussian along with an efficient sampling scheme. We show that the pull-back metric informs distances in the latent space, remaining invariant to reparameterizations. We further make explicit the relationship between uncertainty estimation and proper latent geometry and qualitatively show that geometrically informed priors avoid manifold mismatch by drawing samples from the image of the manifold in the latent space. Quantitatively, we show that our approach outperforms Euclidean VAEs both in an unsupervised learning task and a classification task, especially in low latent space dimensions. 
	
	\subsection*{Acknowledgements}
	SH and DE were supported by a research grant (15334) from VILLUM FONDEN. This project has received funding from the European Research Council (ERC) under the European Union's Horizon 2020 research and innovation programme (grant agreement n\textsuperscript{o} 757360). We gratefully acknowledge the support of the NVIDIA Corporation with the donation of GPU hardware. 
	
	\bibliography{paper}

\begin{thebibliography}{38}
\providecommand{\natexlab}[1]{#1}
\providecommand{\url}[1]{\texttt{#1}}
\expandafter\ifx\csname urlstyle\endcsname\relax
  \providecommand{\doi}[1]{doi: #1}\else
  \providecommand{\doi}{doi: \begingroup \urlstyle{rm}\Url}\fi

\bibitem[Arvanitidis et~al.(2016)Arvanitidis, Hansen, and
  Hauberg]{arvanitidis2016locally}
Arvanitidis, G., Hansen, L.~K., and Hauberg, S.
\newblock A locally adaptive normal distribution.
\newblock In \emph{Advances in Neural Information Processing Systems}, pp.\
  4251--4259, 2016.

\bibitem[Arvanitidis et~al.(2018)Arvanitidis, Hansen, and
  Hauberg]{arvanitidis:iclr:2018}
Arvanitidis, G., Hansen, L.~K., and Hauberg, S.
\newblock Latent space oddity: on the curvature of deep generative models.
\newblock In \emph{International Conference on Learning Representations
  (ICLR)}, 2018.

\bibitem[Arvanitidis et~al.(2019)Arvanitidis, Hauberg, Hennig, and
  Schober]{arvanitidis:aistats:2019}
Arvanitidis, G., Hauberg, S., Hennig, P., and Schober, M.
\newblock Fast and robust shortest paths on manifolds learned from data.
\newblock In \emph{Proceedings of the 19th international Conference on
  Artificial Intelligence and Statistics (AISTATS)}, 2019.

\bibitem[Bauer \& Mnih(2018)Bauer and Mnih]{bauer2018resampled}
Bauer, M. and Mnih, A.
\newblock Resampled priors for variational autoencoders.
\newblock \emph{arXiv preprint arXiv:1810.11428}, 2018.

\bibitem[Bishop(2006)]{Bishop:2006:PRM:1162264}
Bishop, C.~M.
\newblock \emph{{Pattern Recognition and Machine Learning (Information Science
  and Statistics)}}.
\newblock Springer-Verlag New York, Inc., Secaucus, NJ, USA, 2006.

\bibitem[Chen et~al.(2016)Chen, Kingma, Salimans, Duan, Dhariwal, Schulman,
  Sutskever, and Abbeel]{chen2016variational}
Chen, X., Kingma, D.~P., Salimans, T., Duan, Y., Dhariwal, P., Schulman, J.,
  Sutskever, I., and Abbeel, P.
\newblock Variational lossy autoencoder.
\newblock \emph{arXiv preprint arXiv:1611.02731}, 2016.

\bibitem[Davidson et~al.(2018)Davidson, Falorsi, De~Cao, Kipf, and
  Tomczak]{davidson2018hyperspherical}
Davidson, T.~R., Falorsi, L., De~Cao, N., Kipf, T., and Tomczak, J.~M.
\newblock Hyperspherical variational auto-encoders.
\newblock \emph{arXiv preprint arXiv:1804.00891}, 2018.

\bibitem[Detlefsen et~al.(2019)Detlefsen, J{\o}rgensen, and
  Hauberg]{detlefsen2019reliable}
Detlefsen, N.~S., J{\o}rgensen, M., and Hauberg, S.
\newblock Reliable training and estimation of variance networks.
\newblock \emph{arXiv preprint arXiv:1906.03260}, 2019.

\bibitem[do~Carmo(1992)]{docarmo:1992}
do~Carmo, M.
\newblock \emph{Riemannian Geometry}.
\newblock Mathematics (Boston, Mass.). Birkh{\"a}user, 1992.

\bibitem[Eklund \& Hauberg(2019)Eklund and Hauberg]{haubergeklund2019}
Eklund, D. and Hauberg, S.
\newblock Expected path length on random manifolds.
\newblock \emph{arXiv preprint arXiv:1908.07377}, 2019.

\bibitem[Falorsi et~al.(2018)Falorsi, de~Haan, Davidson, De~Cao, Weiler,
  Forr{\'e}, and Cohen]{falorsi2018explorations}
Falorsi, L., de~Haan, P., Davidson, T.~R., De~Cao, N., Weiler, M., Forr{\'e},
  P., and Cohen, T.~S.
\newblock Explorations in homeomorphic variational auto-encoding.
\newblock \emph{arXiv preprint arXiv:1807.04689}, 2018.

\bibitem[Hauberg(2018{\natexlab{a}})]{hauberg2018directional}
Hauberg, S.
\newblock Directional statistics with the spherical normal distribution.
\newblock In \emph{2018 21st International Conference on Information Fusion
  (FUSION)}, pp.\  704--711. IEEE, 2018{\natexlab{a}}.

\bibitem[Hauberg(2018{\natexlab{b}})]{hauberg2018only}
Hauberg, S.
\newblock Only bayes should learn a manifold (on the estimation of differential
  geometric structure from data).
\newblock \emph{arXiv preprint arXiv:1806.04994}, 2018{\natexlab{b}}.

\bibitem[He et~al.(2015)He, Zhang, Ren, and Sun]{DBLP:conf/iccv/HeZRS15}
He, K., Zhang, X., Ren, S., and Sun, J.
\newblock Delving deep into rectifiers: Surpassing human-level performance on
  imagenet classification.
\newblock In \emph{2015 {IEEE} International Conference on Computer Vision,
  {ICCV} 2015, Santiago, Chile, December 7-13, 2015}, pp.\  1026--1034. {IEEE}
  Computer Society, 2015.
\newblock \doi{10.1109/ICCV.2015.123}.
\newblock URL \url{https://doi.org/10.1109/ICCV.2015.123}.

\bibitem[Hennig \& Hauberg(2014)Hennig and Hauberg]{hennig:aistats:2014}
Hennig, P. and Hauberg, S.
\newblock Probabilistic solutions to differential equations and their
  application to riemannian statistics.
\newblock In \emph{Proceedings of the 17th international Conference on
  Artificial Intelligence and Statistics (AISTATS)}, volume~33, 2014.

\bibitem[Hoffman \& Johnson(2016)Hoffman and Johnson]{hoffman2016elbo}
Hoffman, M.~D. and Johnson, M.~J.
\newblock Elbo surgery: yet another way to carve up the variational evidence
  lower bound.
\newblock In \emph{Workshop in Advances in Approximate Bayesian Inference,
  NIPS}, volume~1, pp.\ ~2, 2016.

\bibitem[Hsu(2002)]{hsu2002stochastic}
Hsu, E.~P.
\newblock \emph{Stochastic analysis on manifolds}, volume~38.
\newblock American Mathematical Soc., 2002.

\bibitem[Kingma \& Ba(2015)Kingma and Ba]{DBLP:journals/corr/KingmaB14}
Kingma, D.~P. and Ba, J.
\newblock Adam: {A} method for stochastic optimization.
\newblock In Bengio, Y. and LeCun, Y. (eds.), \emph{3rd International
  Conference on Learning Representations, {ICLR} 2015, San Diego, CA, USA, May
  7-9, 2015, Conference Track Proceedings}, 2015.
\newblock URL \url{http://arxiv.org/abs/1412.6980}.

\bibitem[Kingma \& Welling(2014)Kingma and Welling]{kingma:iclr:2014}
Kingma, D.~P. and Welling, M.
\newblock Auto-{E}ncoding {V}ariational {B}ayes.
\newblock In \emph{Proceedings of the 2nd International Conference on Learning
  Representations (ICLR)}, 2014.

\bibitem[Kingma et~al.(2016)Kingma, Salimans, Jozefowicz, Chen, Sutskever, and
  Welling]{kingma2016improved}
Kingma, D.~P., Salimans, T., Jozefowicz, R., Chen, X., Sutskever, I., and
  Welling, M.
\newblock Improved variational inference with inverse autoregressive flow.
\newblock In \emph{Advances in neural information processing systems}, pp.\
  4743--4751, 2016.

\bibitem[Klushyn et~al.(2019)Klushyn, Chen, Kurle, Cseke, and van~der
  Smagt]{klushyn2019learning}
Klushyn, A., Chen, N., Kurle, R., Cseke, B., and van~der Smagt, P.
\newblock Learning hierarchical priors in vaes.
\newblock In \emph{Advances in Neural Information Processing Systems}, pp.\
  2866--2875, 2019.

\bibitem[Lawrence(2005)]{gplvm}
Lawrence, N.~D.
\newblock Probabilistic non-linear principal component analysis with gaussian
  process latent variable models.
\newblock \emph{Journal of machine learning research}, 6\penalty0
  (Nov):\penalty0 1783--1816, 2005.

\bibitem[Li et~al.(2019)Li, Lindenbaum, Cheng, and
  Cloninger]{li2019variational}
Li, H., Lindenbaum, O., Cheng, X., and Cloninger, A.
\newblock Variational diffusion autoencoders with random walk sampling.
\newblock \emph{arXiv preprint arXiv:1905.12724}, 2019.

\bibitem[Mardia \& Jupp(2000)Mardia and Jupp]{mardia2000basic}
Mardia, K.~V. and Jupp, P.~E.
\newblock Basic concepts and models.
\newblock \emph{Mardia KV, Jupp PE. Directional statistics, 2nd edition.
  Chichester (UK): John Wiley \& Sons}, pp.\  25--56, 2000.

\bibitem[Mathieu et~al.(2019)Mathieu, Le~Lan, Maddison, Tomioka, and
  Teh]{mathieu2019continuous}
Mathieu, E., Le~Lan, C., Maddison, C.~J., Tomioka, R., and Teh, Y.~W.
\newblock Continuous hierarchical representations with poincar{\'e} variational
  auto-encoders.
\newblock In \emph{Advances in neural information processing systems}, pp.\
  12544--12555, 2019.

\bibitem[Nagano et~al.(2019)Nagano, Yamaguchi, Fujita, and
  Koyama]{nagano2019differentiable}
Nagano, Y., Yamaguchi, S., Fujita, Y., and Koyama, M.
\newblock A differentiable gaussian-like distribution on hyperbolic space for
  gradient-based learning.
\newblock \emph{arXiv preprint arXiv:1902.02992}, 2019.

\bibitem[Nalisnick \& Smyth(2016)Nalisnick and Smyth]{nalisnick2016stick}
Nalisnick, E. and Smyth, P.
\newblock Stick-breaking variational autoencoders.
\newblock \emph{arXiv preprint arXiv:1605.06197}, 2016.

\bibitem[Oller(1993)]{oller1993intrinsic}
Oller, J.~M.
\newblock On an intrinsic analysis of statistical estimation.
\newblock In \emph{Multivariate Analysis: Future Directions 2}, pp.\  421--437.
  Elsevier, 1993.

\bibitem[Pennec(2006)]{pennec2006intrinsic}
Pennec, X.
\newblock Intrinsic statistics on riemannian manifolds: Basic tools for
  geometric measurements.
\newblock \emph{Journal of Mathematical Imaging and Vision}, 25\penalty0
  (1):\penalty0 127, 2006.

\bibitem[Que \& Belkin(2016)Que and Belkin]{que:aistats:2016}
Que, Q. and Belkin, M.
\newblock Back to the future: {R}adial basis function networks revisited.
\newblock In \emph{Artificial Intelligence and Statistics (AISTATS)}, 2016.

\bibitem[Rey et~al.(2019)Rey, Menkovski, and Portegies]{rey2019diffusion}
Rey, L. A.~P., Menkovski, V., and Portegies, J.~W.
\newblock Diffusion variational autoencoders.
\newblock \emph{arXiv preprint arXiv:1901.08991}, 2019.

\bibitem[Rezende et~al.(2014)Rezende, Mohamed, and
  Wierstra]{rezende2014stochastic}
Rezende, D.~J., Mohamed, S., and Wierstra, D.
\newblock Stochastic backpropagation and variational inference in deep latent
  gaussian models.
\newblock In \emph{International Conference on Machine Learning}, volume~2,
  2014.

\bibitem[Rosca et~al.(2018)Rosca, Lakshminarayanan, and
  Mohamed]{rosca2018distribution}
Rosca, M., Lakshminarayanan, B., and Mohamed, S.
\newblock Distribution matching in variational inference.
\newblock \emph{arXiv preprint arXiv:1802.06847}, 2018.

\bibitem[Skopek et~al.(2019)Skopek, Ganea, and B{\'e}cigneul]{skopek2019mixed}
Skopek, O., Ganea, O.-E., and B{\'e}cigneul, G.
\newblock Mixed-curvature variational autoencoders.
\newblock \emph{arXiv preprint arXiv:1911.08411}, 2019.

\bibitem[Tomczak \& Welling(2017)Tomczak and Welling]{tomczak2017vae}
Tomczak, J.~M. and Welling, M.
\newblock Vae with a vampprior.
\newblock \emph{arXiv preprint arXiv:1705.07120}, 2017.

\bibitem[Tosi et~al.(2014)Tosi, Hauberg, Vellido, and Lawrence]{Tosi:UAI:2014}
Tosi, A., Hauberg, S., Vellido, A., and Lawrence, N.~D.
\newblock {Metrics for Probabilistic Geometries}.
\newblock In \emph{The Conference on Uncertainty in Artificial Intelligence
  (UAI)}, July 2014.

\bibitem[van~den Oord et~al.(2017)van~den Oord, Vinyals, et~al.]{van2017neural}
van~den Oord, A., Vinyals, O., et~al.
\newblock Neural discrete representation learning.
\newblock In \emph{Advances in Neural Information Processing Systems}, pp.\
  6306--6315, 2017.

\bibitem[Xu \& Durrett(2018)Xu and Durrett]{xu2018spherical}
Xu, J. and Durrett, G.
\newblock Spherical latent spaces for stable variational autoencoders.
\newblock \emph{arXiv preprint arXiv:1808.10805}, 2018.

\end{thebibliography}
	\bibliographystyle{icml2020}
	\newpage
	\onecolumn
	\appendix
	\section{On neural network-based immersions}
	For the decoder map \ref{eq:f-map} to be a valid immersion, its differential $\dif f$ needs to be injective for all $p \in \mathcal{M}$ as stated in definition \ref{def:immersions}. The differential of $f$ is represented by its Jacobian matrix $\mat{J_f}$ and for it to be injective for all $p \in \mathcal{M}$, it needs to be full rank. This is ensured if for the MLPs representing the decoder $\boldsymbol{\mu}_\theta$ and $\boldsymbol{\sigma}_\psi$ the following are true:
	\begin{itemize}
		\item Each hidden layer in the network has an equal or greater number of units to the previous layer ($n_{L - 1} \leq n_{L}$).
		\item All weight matrices in the network are full rank.
		\item The activation functions are at least twice differentiable and strictly monotone.
	\end{itemize}
	
	In our experiments, we opt for the same number of units in each hidden layer of the network and ELU non-linearities. In theory, the ELU activation function could present problems since it has a point of discontinuity at 0, however we did not experience any numerical instability that would arise in such case. All weight matrices are initialized uniformly \cite{DBLP:conf/iccv/HeZRS15} which practically has zero probability of yielding low rank weight matrices. While theoretically this could change via the gradient updates of the weights, this would once again immediately break experiments because of numerical instabilities, which we did not observe.
	
	\section{Geodesic estimation}
	We estimate geodesic distances by minimizing curve energy. In detail, we represent the geodesic curve with a cubic spline with parameters initialized to form a straight line. These parameters are then optimized via gradient descent by minimizing the curve energy: 
	\begin{align}
	\mathcal{E}(\boldsymbol{\gamma}) &= \frac{1}{2} \int_{0}^{1}||\boldsymbol{\dot{\gamma}}(t)||_{g}^{2}\, \dif t \nonumber \\
	&= \frac{1}{2} \int_{0}^{1} \boldsymbol{\dot{\gamma}}^{T}(t) \mat{G}_{\boldsymbol{\gamma}} \boldsymbol{\dot{\gamma}}(t)\, \dif t \label{app:eq_energy}
	\end{align}
	where $\boldsymbol{\gamma}$ is the geodesic curve, $\boldsymbol{\dot{\gamma}}$ is the first derivative of the curve, i.e. its velocity vector and $\mat{G}_{\boldsymbol{\gamma}}$ is the matrix representation of the metric tensor evaluated at the curve points. The integral \ref{app:eq_energy} is computed by numerical approximation, where the partition of the interval can be chosen as a hyperparameter.
	
	\section{Experimental setup}
	The architectures of all model variants are shown below in Tables \ref{app_tab:encoder} and \ref{app_tab:decoder}. The encoder mean and variance, as well as the decoder mean are modelled by 2-layer MLPs as shown below. The decoder mean mirrors the encoder mean, while the \emph{precision} $\boldsymbol{\beta}$ is estimated by the RBF network. The number of the RBF centers is set to 350 and the bandwidth is set to 0.01 in all cases. For a fair comparison, all models share the same underlying architecture for the encoder and decoder. Tables \ref{app_tab:encoder} and \ref{app_tab:decoder} summarize the architectures, listing the activation function for each layer with the units corresponding to each layer in parentheses.
	
	\begin{table}[h]
		\centering
		\caption{Encoder network architectures.}
		\begin{tabular}{cccc}
			Network & Layer 1 & Layer 2 & Output \\
			\toprule
			$\boldsymbol{\mu_\phi(\x)}$ & ELU (300) & ELU (300) & Linear (\textit{dim($\Z$)}) \\
			$\boldsymbol{\sigma^2_\phi(\x)}$ & ELU (300) & ELU (300) & Softplus (\textit{dim($\Z$)})
		\end{tabular}
		\label{app_tab:encoder}
	\end{table}
	
	\begin{table}[h]
		\centering
		\caption{Decoder network architectures. * denotes strictly positive weights.}
		\begin{tabular}{cccc}
			Network & Layer 1 & Layer 2 & Output \\
			\toprule
			$\boldsymbol{\mu_\theta(\z)}$ & ELU (300) & ELU (300) & Linear (\textit{dim($\X$)}) \\
			$\boldsymbol{\beta_\psi(\z)}$ & RBF ($\mathbb{R}^{dim(\mathcal{Z}) \times 350}$) & Linear* (\textit{dim($\X$)}) & Identity (\textit{dim($\X$}))
		\end{tabular}
		\label{app_tab:decoder}
	\end{table}
	
	\subsection{Section 5.1 experiment}
	\citet{detlefsen2019reliable} highlighted the importance of optimizing the mean and variance components separately, when training VAEs with Gaussian generative models. Following this paradigm, in all our experiments we first optimize the encoder components ($\boldsymbol{\mu_\phi}$ and $\boldsymbol{\sigma_\phi}$) along with the decoder $\boldsymbol{\mu_\theta}$. Then, keeping these fixed, we optimize the decoder $\boldsymbol{\sigma_\psi}$. All models were trained for 300 epochs. More specifically, the $\mathcal{R}$-VAE was trained as an autoencoder (optimizing only the encoder $\boldsymbol{\mu_\phi}$ and $\boldsymbol{\sigma_\phi}$ and the decoder $\boldsymbol{\mu_\theta}$) for the first 100 epochs and for the remaining 200 epochs the latent prior and the decoder $\boldsymbol{\beta_\psi}$ were optimized. Similarly for a VAE, it was deterministically warmed up for 100 epochs and for the remaining 200 epochs, the decoder $\boldsymbol{\beta_\psi}$ was optimized. All experiments were run with the Adam optimizer \cite{DBLP:journals/corr/KingmaB14} with default parameter settings and a fixed learning rate of $10^{-3}$. The batch size was 100 for all models.
	
	\subsection{Section 5.2 experiment}
	The classifier used on this section was a single, 100-unit layer MLP with ReLU non-linearities, trained for 100 epochs with the Adam optimizer with default parameter settings and a learning rate of $10^{-3}$. The batch size was set at 64. The architectures of the models giving rise to the latent representations are as in the previous section.
	
	\subsection{Runtime comparisons}
	Below is the wall clock time for every model used in the experiments. The statistics were computed without a fixed seed. The latent space dimensions are denoted by $d$. In VAE-VampPrior, $n$ denotes the number of mixture components in the latent prior.
	\begin{table}[h]
		\centering
		\caption{Per epoch training time for each model. Mean and std deviation in seconds, computed over 100 epochs on MNIST.}
		\begin{tabular}{l|ccc}
			Model & $d=2$ & $d=5$ & $d=10$ \\
			\hline
			VAE & 10.64$_{\pm.51}$ & 11.01$_{\pm.77}$ & 11.10$_{\pm.60}$ \\
			VAE-VampPrior ($n=128$) & 10.66$_{\pm.6}$ & 11.22$_{\pm.9}$ & 11.37$_{\pm.96}$ \\
			VAE-VampPrior ($n=256$) & 10.72$_{\pm.3}$ & 11.34$_{\pm1.21}$ & 11.52$_{\pm.77}$ \\
			VAE-VampPrior ($n=512$) & 10.9$_{\pm.34}$ & 11.38$_{\pm.93}$ & 12.18$_{\pm1.12}$\\
			$\mathcal{R}$-VAE & 55.73$_{\pm4.36}$ & 59.97$_{\pm1.33}$ & 60.13$_{\pm1.19}$ 
		\end{tabular}
		\label{app_tab:runtime}
	\end{table}
	
	\subsection{Complete results for VAE-VampPrior}
	Tables~\ref{tab:mnist} \& \ref{tab:fmnist} show the results of the best performing VampPrior model variant. Here we show the complete results of the VAE-VampPrior in all settings. Below $n$ denotes the number of mixture components in the latent prior, while $d$ denotes the latent space dimensions.
	\begin{table}[h!]
		\centering
		\caption{MNIST results of VAE with VampPrior for varying latent space dimensions and number of mixture components in the latent prior.}
		\begin{tabular}{lccc}
			Model & Neg. ELBO & Rec & KL \\ 
			\toprule
			\multicolumn{4}{c}{$d=2$} \\
			\hline
			VAE-VampPrior ($n=128$) & -1039.66$_{\pm2.56}$ & -1042.13$_{\pm2.56}$  & 2.46$_{\pm.01}$ \\
			VAE-VampPrior ($n=256$) & -1045.04$_{\pm5.20}$ & -1047.34$_{\pm5.22}$  & 2.30$_{\pm.03}$ \\
			VAE-VampPrior ($n=512$) & -1040.79$_{\pm9.23}$ & -1043.24$_{\pm9.25}$ & 2.45$_{\pm.05}$ \\
			\hline
			\multicolumn{4}{c}{$d=5$} \\
			\hline
			VAE-VampPrior ($n=128$) & -1100.77$_{\pm4.98}$ & -1102.46$_{\pm4.91}$ & 1.69$_{\pm.06}$ \\
			VAE-VampPrior ($n=256$) & -1103.29$_{\pm1.85}$ & -1105.04$_{\pm1.79}$ & 1.75$_{\pm.12}$ \\
			VAE-VampPrior ($n=512$) & -1109.74$_{\pm4.87}$ & -1111.63$_{\pm4.87}$ & 1.88$_{\pm.01}$ \\
			\hline
			\multicolumn{4}{c}{$d=10$} \\
			\hline
			VAE-VampPrior ($n=128$) & -1110.05$_{\pm6.10}$ & -1112.23$_{\pm5.82}$ & 1.84$_{\pm.04}$ \\
			VAE-VampPrior ($n=256$) & -1116.58$_{\pm4.23}$ & -1118.28$_{\pm4.20}$ & 1.69$_{\pm.02}$\\
			VAE-VampPrior ($n=512$) & -1100.64$_{\pm2.93}$ & -1102.42$_{\pm2.97}$ & 1.78$_{\pm.03}$ \\
		\end{tabular}
	\end{table}
	
	\begin{table}
		\centering
		\caption{FashionMNIST results of VAE with VampPrior for varying latent space dimensions and number of mixture components in the latent prior.}
		\begin{tabular}{lccc}
			Model & Neg. ELBO & Rec & KL \\ 
			\toprule
			\multicolumn{4}{c}{$d=2$} \\
			\hline
			VAE-VampPrior ($n=128$) & -694.63$_{\pm8.65}$ & -697.14$_{\pm8.65}$  & 2.50$_{\pm.01}$ \\
			VAE-VampPrior ($n=256$) & -702.67$_{\pm17.45}$ & -705.19$_{\pm17.44}$  & 2.52$_{\pm.04}$ \\
			VAE-VampPrior ($n=512$) & -705.90$_{\pm21.29}$ & -708.45$_{\pm21.29}$ & 2.54$_{\pm.01}$ \\
			\hline
			\multicolumn{4}{c}{$d=5$} \\
			\hline
			VAE-VampPrior ($n=128$) & -755.80$_{\pm.66}$ & -756.58$_{\pm.71}$ & 0.77$_{\pm.06}$ \\
			VAE-VampPrior ($n=256$) & -767.54$_{\pm3.22}$ & -768.33$_{\pm3.31}$ & 0.78$_{\pm.09}$ \\
			VAE-VampPrior ($n=512$) & -769.27$_{\pm5.0}$ & -770.10$_{\pm5.02}$ & 0.83$_{\pm.09}$ \\
			\hline
			\multicolumn{4}{c}{$d=10$} \\
			\hline
			VAE-VampPrior ($n=128$) & -754.47$_{\pm6.78}$ & -758.20$_{\pm6.72}$ & 3.72$_{\pm.06}$ \\
			VAE-VampPrior ($n=256$) & -756.13$_{\pm5.40}$ & -760.49$_{\pm5.1}$ & 3.69$_{\pm.06}$\\
			VAE-VampPrior ($n=512$) & -774.17$_{\pm10.83}$ & -777.75$_{\pm10.78}$ & 3.58$_{\pm.06}$ \\
		\end{tabular}
	\end{table}
\end{document}